\documentclass[ sn-mathphys-num, Numbered]{sn-jnl}
\usepackage{graphicx}
\usepackage{multirow}
\usepackage{amsmath,amssymb,amsfonts}
\usepackage{amsthm}
\usepackage{mathrsfs}
\usepackage[title]{appendix}
\usepackage{xcolor}
\usepackage{textcomp}
\usepackage{manyfoot}
\usepackage{booktabs}
\usepackage{algorithm}
\usepackage{algorithmicx}
\usepackage{algpseudocode}
\usepackage{listings}
\usepackage{subcaption}
\usepackage{graphicx}
\usepackage{colortbl}
\usepackage{tabulary}
\usepackage{etoolbox}
\usepackage{tabularx,booktabs}
\usepackage{amssymb}
\usepackage{commath}
\usepackage{mathtools}
\usepackage{IEEEtrantools}
\usepackage{bm}
\usepackage{upgreek}
\usepackage{dsfont}

\usepackage{graphicx}        
\usepackage{xcolor}
\usepackage{transparent}
\usepackage[percent]{overpic}
\usepackage{pgfplots}
\usepackage{multirow}
\usepackage{multicol}
\usepackage{makecell}
\usepackage{tabularx}
\usepackage{booktabs}
\usepackage{caption}         
\usepackage{subcaption}
\usepackage{paralist}
\usepackage[shortlabels]{enumitem}
\usepackage{times}
\usepackage{textcomp}        
\usepackage{gensymb}
\usepackage[normalem]{ulem}
\usepackage[htt]{hyphenat}
\usepackage{csquotes}
\usepackage{comment}
\usepackage[binary-units]{siunitx}
\usepackage{xurl}
\usepackage{nameref}
\usepackage{hyperref}
\usepackage{import}

\date{}
\newcommand\blfootnote[1]{
    \bgroup
    \renewcommand\thefootnote{\fnsymbol{footnote}}
    \renewcommand\thempfootnote{\fnsymbol{mpfootnote}}
    \footnotetext[0]{#1}
    \egroup
}

\theoremstyle{thmstyleone}

\theoremstyle{thmstyletwo}

\theoremstyle{thmstylethree}

\begin{document}
\title[Article Title]{Towards Shared Embodied Intelligence in Humanoid Robots through Optimization Development and Testing of the Human-Aware ergoCub Robot}
\author*[1,2,3]{\fnm{Carlotta} \sur{Sartore}}\email{carlotta.sartore@gbionics.ai}
\author[2]{\fnm{Mohamed} \sur{Elobaid}}
\author[1,2]{\fnm{Lorenzo} \sur{Rapetti}}
\author[1,2]{\fnm{Giulio} \sur{Romuladi}}
\author[1,2]{\fnm{Stefano} \sur{Dafarra}}
\author[4]{\fnm{Nicola A.} \sur{Piga}}
\author[1,2,3]{\fnm{Ines} \sur{Sorrentino}}
\author[1,2]{\fnm{Paolo Maria} \sur{Viceconte}}
\author[1,2]{\fnm{Silvio} \sur{Traversaro}}
\author[5]{\fnm{Ugo} \sur{Pattacini}}
\author[1,5]{\fnm{Luca} \sur{Fiorio}}
\author[6]{\fnm{Francesco} \sur{Draicchio}}
\author[6]{\fnm{Giovanna} \sur{Tranfo}}
\author[4]{\fnm{Lorenzo} \sur{Natale}}
\author[1,5]{\fnm{Marco} \sur{Maggiali}}
\author*[1,2]{\fnm{Daniele} \sur{Pucci}}\email{daniele.pucci@gbionics.ai}
\affil[1]{\orgdiv{} \orgname{GenerativeBionics}, \orgaddress{\city{Genova}, \country{Italy}}}
\affil[2]{\orgdiv{Artificial and Mechanical Intelligence}, \orgname{Istituto Italiano di Tecnologia}, \orgaddress{\city{Genova}, \country{Italy}}}
\affil[3]{\orgdiv{School of Computer Science}, \orgname{University of Manchester}, \orgaddress{\city{Manchester}, \country{United Kingdom}}}
\affil[4]{\orgdiv{Humanoid Sensing and Perception}, \orgname{Istituto Italiano di Tecnologia}, \orgaddress{\city{Genova}, \country{Italy}}}
\affil[5]{\orgdiv{iCub Tech Facility}, \orgname{Istituto Italiano di Tecnologia}, \orgaddress{\city{Genova}, \country{Italy}}}
\affil[6]{\orgdiv{DiMEILA}, \orgname{Istituto Nazionale Assicurazione Infortuni sul Lavoro (INAIL)}, \orgaddress{\city{Rome}, \country{Italy}}}
\abstract{Collaboration is central to human behavior, enabling tasks beyond individual capability. This ability arises from coordinating actions through internal representations of others, a concept known as \emph{shared intelligence}. Additionally, humans are characterized by physical bodies and cognitive abilities that are optimized in response to their environment, a phenomenon referred to as \emph{embodied cognition}. Designing humanoid robots that collaborate safely and effectively with people requires unifying these principles. Here we propose an architecture that integrates \emph{shared intelligence} and \emph{embodied cognition} to enable robots to physically collaborate with humans, where robot hardware and control are optimized for human metrics, using representations of the human body and motion intelligence. The ultimate goal is to achieve a form of \emph{shared embodied intelligence}. Specifically, our architecture optimizes robot hardware and physical intelligence parameters with respect to human ergonomic metrics. This is accomplished by modeling human–robot interaction as a function of hardware configurations and embedding human models into the robot’s physical intelligence. As a concrete implementation, we present the humanoid robot ergoCub, whose morphology and control have been optimized for collaborative tasks with humans. Our approach provides a framework for designing humanoid robots that prioritize human ergonomics at both the hardware and physical intelligence levels, with applications in industrial and assistive robotics.}
\maketitle
\graphicspath{{Introduction/Figures/}} 
\section{Introduction}
\label{sec:Introduction}
\emph{Collaboration} is a conducive condition for survival in the natural world, enabling groups to accomplish tasks unattainable by individuals. For instance, wolves leverage complex social dynamics to ensure both the pack's and the individual's survival \cite{cordoni2019back}; ants coordinate to transport heavy loads; and human orchestras produce sounds impossible for individuals to create alone. Collaboration can even overcome species limitations, leading to interspecies partnerships such as the dog-human alliance, where dogs' acute sense of smell complements humans' strategic planning, resulting in highly effective rescue missions \cite{fenton1992use}. It becomes then clear how collaboration depends upon a degree of human \emph{shared intelligence}, namely, the capability to perform coordinated actions and maintain mutual awareness, thereby fostering intelligent behavior that surpasses individual capacities.
An individual's intelligence is also shaped by physical characteristics. This is evident in evolution, where physical traits develop alongside cognition, forming what is known as embodied cognition \cite{morris2003life, wilson2011embodied, wilson2013embodied}. For example, consider the human ability to walk, which results from the synergy and coordination of muscle reflexes and neural circuits controlling gait \cite{Geyer2010}. Physical attributes have also evolved in response to the need for collaboration within and across species. An example is the oxpecker bird, whose beak has evolved to precisely remove parasites from its hosts without causing harm \cite{nunn2011mutualism}.
Inspired by the above observations, this paper lays the foundation for \emph{shared embodied intelligence} architectures for artificial agents, and more precisely for humanoid robots.   By \emph{shared embodied intelligence}, we refer to the integration of \emph{shared intelligence} and \emph{embodied cognition} to equip an agent with both physical attributes and cognitive abilities optimized for collaboration with other beings. The proposed shared embodied intelligence architecture enables simultaneous optimization of the robot's physical intelligence and hardware characteristics, taking into account not only the task and environment but also the physical intelligence and attributes of potential partners, whether human or robotic. We focus on humanoid robots as the target platform for our shared embodied intelligence architecture, since their human-like form enables more natural physical interaction and facilitates task execution in environments designed for humans \cite{gu2025humanoid}.
The presented shared embodied intelligence principles are applied to design a humanoid robot: the ergoCub (Figure \ref{fig:ergocub}), developed at the Istituto Italiano di Tecnologia (IIT) in collaboration with the National Institute for Insurance against Accidents at Work (INAIL).
\subsection{Related works}
In the literature, physical intelligence has been instantiated in various forms, including behavior models \cite{trilbmteam2025carefulexaminationlargebehavior}, imitation-learning policies \cite{cardenas2024xbg}, and foundation VLA models \cite{nvidia2025gr00tn1openfoundation}. A widely adopted approach for humanoid robots is the cascade control architecture \cite{feng2015optimization}. This architecture is structured into multiple layers - or control loops - each with a different role and degree of complexity. The outer layers provide references to achieve the desired goal while the inner layers of the control architecture receive inputs from the robot and from the environment, and they provide references to subsequent loops.
The further down a control loop is within the hierarchical architecture, the more complex the robot models become and the shorter the prediction horizons are for generating outputs, which often enables each inner loop to run faster than the outer ones.
 Specifically, the standard \emph{physical intelligence} control architecture for humanoid robots comprises three fundamental layers: the \emph{trajectory generation}, the \emph{trajectory adjustment}, and the \emph{trajectory control} layer. The \emph{trajectory generation} layer translates high-level objectives into whole-body reference trajectories, typically using direct of indirect trajectory optimization or data-driven methods, often computed off-line without robot feedback \cite{Viceconte2022,Dafarra2022,herzog2015trajectory}. The \emph{trajectory adjustment} layer refines these trajectories using simplified models and on-line feedback \cite{Romualdi2022,griffin2016model,joe2018balance}, which are then executed by the \emph{trajectory control} layer via inputs such as currents or voltages, commonly through Quadratic Programming (QP) \cite{ott2011, wensing2013, Nava2016}. This hierarchical control has also been applied to human–robot collaboration, with emphasis on safety, efficiency \cite{bauer2008,haddadin2016}, and \emph{ergonomics} \cite{rasmussen2002design, lagomarsino2022} to align with human physical and cognitive abilities.
Despite the above efforts, shared actions remain elusive due to limited partner representation, leading to insufficient mutual partner awareness. 
When attempting energy efficiency or increasing robot performances, \emph{physical intelligence} may not be enough to achieve the desired requirements. \emph{Robot hardware} plays a pivotal role in these cases.
However, the properties of the robot hardware are sometimes overlooked during the development of control architectures for implementing robot \emph{physical intelligence}. Robot hardware is often treated as a given, immutable component. Conversely, during robot hardware design, the specific task the robot is intended to perform is sometimes disregarded, leading to suboptimal designs that do not align with the robot's intended functions.
To optimize robot hardware and control simultaneously, the principle of \emph{co-design}  has been largely applied to the design process of robots. Co-design places a strong emphasis on optimizing parameters characterizing both the robot hardware and the control system to obtain optimal body-and-control robot concepts. For instance, the Alternating Direction Method of Multipliers (ADMM) has been used to identify quadruped hardware parameters optimized for walking on different terrains 
\cite{bravo2022large}. Additionally, reinforcement learning has been applied to the co-design problem considering the hardware as part of the policy showing evidence of how a task directly affects the design of the robot \cite{ha2019reinforcement,Bjelonic2023}. Tackling the same problem, genetic algorithms have been used to identify optimum control and hardware able to ensure control robustness while minimizing energy consumption \cite{ fadini2022simulation, Gkikakis2021,bergonti2023codesign}. 
However, these studies focus on robots composed of a limited number of links, overlooking the inherent complexities of humanoid robots. Moreover, co-design principles are applied to individual robot subparts rather than considering the entire robot, missing the synergistic interplay among different robot components. Furthermore, co-design principles are yet to be applied to tailor robot designs addressing physical human-robot collaboration tasks.
  Our previous works \cite{Sartore2022, sartore2023codesign} presented the utilization of both classical optimization techniques and genetic algorithms to co-design humanoid robots for human-robot collaborative tasks still not considering human simulation and identifying hardware parameters under static conditions. Furthermore , they neither address the translation of the optimization output into real robot hardware, nor the integration of the optimal hardware with robot physical intelligence.
\subsection{Contribution}
This paper proposes a methodology to optimize robot hardware and control parameters considering not only the robot body, environment, and task - as classically attempted when implementing instances of optimal \emph{embodied cognition} - but also possible partners possessing themselves a degree of \emph{physical intelligence}. We see this as a first step towards artificial agents owning a degree of \emph{shared embodied intelligence}, namely having humanoid robots designed considering the body and properties of surrounding individuals. More specifically, the contribution is two-fold:
\begin{enumerate}
    \item \emph{Scientific}: we present a \emph{shared embodied intelligence} architecture that can be employed to optimize both the hardware and control of humanoid robots interacting physically with humans to achieve specific tasks. The key ingredients of the architecture are: $i)$ differentiable parametrized floating-base models representing humanoid robots and humans that exchange contact forces for achieving a physical collaboration task; the parameters of the models are robot limbs lengths and ensure physical consistency of the underlying mechanics (e.g. inertia) $ii)$ a parametrized physical intelligence architecture for the humanoid robot that considers a representation of the physical intelligence of the human, which is implemented by modeling the policies that generate human movements, namely, signals that affect the human musculoskeletal system. The parameters of this architecture are those that characterize how the robot physical intelligence is implemented (e.g. gains of the low-level controllers, horizons of the high-level planners, etc.). The architecture, the controllers for the robot physical intelligence, and the differentiable models are publicly available in Python-based libraries (\url{https://github.com/ami-iit/adam}, \url{https://github.com/ami-iit/shared-controllers}). 
    \item \emph{Technological}: we present the ergoCub humanoid robot, which is designed and controlled using the above architecture for shared embodied intelligence. The name of the humanoid robot is composed of two words: \emph{ergo}, which stands for \emph{ergonomics}, and \emph{Cub}, which references the \emph{iCub} humanoid robot that represented a starting point for the optimizations performed for the ergoCub. The emphasis on ergonomics is due to the fact that the robot was optimized with respect to two tasks: walking and collaborating with humans during payload lifting tasks. More precisely, the parameter optimization layer of the architecture considers quantitative human ergonomic indices, specifically human back stress, to identify parameters characterizing both the robot hardware and its physical intelligence, while simultaneously improving locomotion robustness and minimizing energy consumption. These metrics capture instantaneous biomechanical risk during human–robot interaction and therefore quantify musculoskeletal loading. As such, they do not necessarily align with subjective human measures such as perceived comfort or fatigue.
\end{enumerate}
The ergoCub robot, coupled with its cognitive architecture, represents the inaugural embodiment of our methodology. 
\graphicspath{{Figures/}} 
\section{Results}
\label{sec:Results}
Shared embodied intelligence enables collaborative tasks by jointly optimizing robot hardware and control parameters with respect to ergonomic indexes. We designed a modular architecture based on human and robot representations to identify optimal hardware and physical intelligence parameters for human–robot collaboration. Here, \emph{physical intelligence} refers to the components that enable behaviors through movements, including trajectory blocks, sensors/sensory systems, and muscle/actuators (see Section~\ref{sec:material:modelling}).
To design and control the ergoCub humanoid, we implemented two versions of this architecture. The first optimizes hardware by adjusting link lengths to improve walking performances and ergonomics during static collaborations (Figure~\ref{fig:istances}a), assuming simplified link geometries to guide manufacturing. The second (Figure \ref{fig:istances}b) takes the optimized hardware as input and is aimed to improve human-robot coordination and robustness during task execution by optimizing the robot's physical intelligence parameters, like gains and weights of the optimal control problems.
Together, these two instances culminate in the ergoCub robot, which integrates optimized hardware and control. The following sections present and assess their results, validating our shared embodied intelligence approach.
\subsection{First architecture instance: ergoCub optimal hardware}
\label{sec:results:ergocubRobot}
The ergoCub robot is the outcome of a concatenation of manufacturing and optimization processes, as visually represented in Figure \ref{fig:istances}a.
The hardware optimization employed non-linear techniques, which are highly sensitive to initial conditions. To address this, we used the iCub3 humanoid as a starting point, retaining its kinematic structure while optimizing link geometries for selected tasks. iCub3 is an open-source humanoid robot developed at IIT and widely adopted in applications from locomotion to telepresence \cite{dafarra2024icub3}, providing a solid baseline that ensured the optimization remained both effective and efficient.
To maintain the iCub3 kinematic structure, electronics, and type of motors we imposed a constraint on the ergoCub maximum height to ensure manufacturability. 
In detail, we start by representing the iCub3 humanoid robot links using basic geometric shapes such as spheres, cylinders, and boxes to create a simplified model, depicted in Figure \ref{fig:istances}a. This approach allowed us to generalize the dynamic characteristics of the robot's links based on their lengths. Subsequently, we analyze the influence of the robot hardware parameters on human-robot interaction by leveraging representations of the human body and physical intelligence. 
The key components of the human models are depicted in Figure \ref{fig:istances}a and they consist of policies that generate human motions, namely, signals that affect the human musculoskeletal system. Specifically, we modeled the human control system by decomposing it into two main components: a centralized processing unit representing the central nervous system's role, and a local processing unit representing the output of the peripheral nervous system. The centralized processing unit models the high-level motion commands, and it includes whole-body control and trajectory blocks. The model of the peripheral nervous system concerns the joint control loop, and it includes blocks for gathering sensory information and stabilizing motion through localized muscle reflexes. 
Further details are provided in Section \ref{sec:methods:robot-load-coupled-dynamics}.
Starting from the defined human and robot model, we formulated the human-robot coupled dynamics, parametrized with respect to the robot hardware parameters. Such a coupled human-robot parametrized dynamic is harnessed to define an optimization problem to find the robot optimum link lengths. In this optimization, our focus is on a scenario where a human and a robot collaborate to lift a load. More precisely, our objective is two-fold : enhancing walking capabilities and improving the ergonomics of human-robot physical interaction. In this way, we will design a humanoid robot able to collaborate with humans in lifting a load but we will also consider additional humanoid robot key abilities such as walking.
More in detail, the optimization problem was defined with two main objectives. The first was to improve ergonomics during human–robot collaborative lifting by minimizing the energy expenditure of both agents, quantified as joint torques, across several static lifting configurations (see Section~\ref{sec:methods:morphology_optimization}). The second objective was to enhance walking performance by raising the robot’s center of mass, thereby increasing system bandwidth and robustness \cite{Choi2006}.
\noindent More detail about the formulation and mathematical background of the optimization problem defined can be found in Section \ref{sec:methods:morphology_optimization}.
In Figure \ref{fig:hardware_optimization_output}, we present the outputs of the hardware optimization process. Figure \ref{fig:hardware_optimization_output} (a) shows the simplified model of the iCub3 robot,    depicted in red, serving as the initial configuration for our optimization. Figures \ref{fig:hardware_optimization_output} (b) and \ref{fig:hardware_optimization_output} (c) depict the robot designs obtained through the proposed optimization, considering load height ranges of \SIrange{0.8}{1.2}{\meter} and \SIrange{0.8}{1.5}{\meter}, respectively.   We will refer to these models as the optimized blue model and the optimized green model respectively. 
Figure \ref{fig:hardware_optimization_output} (d) summarizes the output of the performed optimization by reporting the human joint torque computed for the original, optimized blue, and optimized green robot designs. These torque values are deterministic and directly tied to the cost function of the optimization problem (more detail can be found in Section \ref{sec:methods:morphology_optimization}). Real-world testing involving human-robot physical collaborations is presented separately in Section \ref{sec:results:ergoCub_physical_intelligence}.
Compared to the original design (Figure~\ref{fig:hardware_optimization_output} a), the optimized blue model (Figure~\ref{fig:hardware_optimization_output} (b)) reduces human back stress at all load heights, with the greatest relief observed at the lumbosacral joint (L5–S1), a region particularly vulnerable to high compressive forces during lifting \cite{waters1993revised}. The green model (Figure~\ref{fig:hardware_optimization_output} (c)) further decreases back torque at nearly all heights, except at \SI{0.8}{\meter}, and uniquely enables lifting tasks up to \SI{1.5}{\meter}. For these reasons, we selected the green solution, as it provides the best compromise between extended task reach and improved ergonomics, outperforming both the original design (red in Figure~\ref{fig:hardware_optimization_output} a) and the other optimized candidate (blue in Figure~\ref{fig:hardware_optimization_output} (b)).
The limb lengths of the selected solution are provided in Figure \ref{fig:hardware_optimization_output} (e). Starting with the identified optimal limb lengths, we built ergoCub and  employed metal extenders to achieve the desired measurements.
In the same Figure \ref{fig:hardware_optimization_output} (e), we report the main dimensions of ergoCub compared with iCub3, and in Figure~\ref{fig:ergocub}a we illustrate additional features of ergoCub. Further analyses are provided in the supplementary material, including a comparison of collaborative payload lifting between ergoCub and iCub3 (Appendix B). To enhance human–robot coordination, ergoCub is equipped with a flexible LCD screen that displays simplified facial expressions, enabling feedback during task execution (see Section~\ref{sec:results:ergoCub_physical_intelligence}). The acceptability of the ergoCUb design was evaluated in a survey with 850 participants from industrial and healthcare domains. ergoCub was rated more positively than Baxter \cite{Baxter} and R1 \cite{R12017}, and its simplified facial expressions were consistently preferred (Appendix E) over the alternatives.
\subsection{Second architecture instance: ergoCub physical intelligence}
\label{sec:results:ergoCub_physical_intelligence}
In this section, we examine the ergoCub’s physical intelligence, namely its ability to generate movements that achieve the desired behavior. Building on the optimized hardware described in Section~\ref{sec:results:ergocubRobot}, we refined the control architecture shown in Figure~\ref{fig:istances}b to address both collaboration and locomotion. For collaboration, the framework mirrors human and robot physical intelligence as hierarchical systems composed of trajectory blocks, sensors, and actuators. Such symmetry allows the robot to account for the human partner's behavior and identify strategies that enhance collaboration. For locomotion, the human component is disregarded, and the framework focuses on walking performance. We present both collaboration and locomotion instances since the hardware optimization enhanced both tasks. Appendix A describes an additional validation scenario in which a human and ergoCub walk together while carrying a payload.
\subsubsection*{Human-robot collaboration tasks}
We validated the optimized hardware and shared embodied intelligence architecture for physical intelligence (Figure \ref{fig:istances}b) on the real ergoCub robot. In the tested scenario (Figure \ref{fig:ergocub}b–d), the robot collaborates with a human partner to manipulate payloads of varying weights: empty, \SI{1}{\kilogram}, and \SI{2}{\kilogram}. During these collaborations, the robot follows the human’s upward and downward movements while the human’s stress is monitored. These experiments are conducted in a dynamic setting, with the human and robot jointly moving the load, thereby relaxing the static assumptions of the hardware optimization.
In an initial offline phase, a human model is initialized by measuring the subject’s height and weight, following \cite{latella2019simultaneous}. During collaboration, this model is continuously updated by an estimation algorithm using data from non-intrusive wearable sensors (Appendix E). More specifically, human kinematics are estimated with an inverse kinematics approach \cite{rapetti2020model}, while joint torques are computed using a Maximum-A-Posteriori method \cite{latella2019simultaneous,tirupachuri2021online} and smoothed with a moving average filter (window size = 2). These estimates continuously update the internal human model of the shared embodied intelligence architecture.
Thanks to this approach, the robot’s physical intelligence can adapt to different human partners, capturing individual physical characteristics through offline initialization of the human model, and online updates of the human kinematics and dynamics via sensor feedback. 
The robot's reference trajectories are pre-planned to optimize the configuration during human interaction, while the robot's movement velocity (in terms of both amplitude and direction) is governed by a variable that depends on the height difference between the robot's and the human's hands. The theoretical framework of this architecture is detailed in Section \ref{sec:method:human-robot-coll}.
During task execution, the robot continuously monitors the human's ergonomic state by estimating the torque exerted on the lumbosacral joint (L5-S1), a critical indicator of ergonomic strain in lifting tasks. Based on the detected stress levels at the L5-S1 joint, the robot uses its head-mounted LCD screen to display expressive feedback, signaling whether the human is experiencing high or low stress.
Moreover, the robot follows the human only within a predefined height range of \SIrange{0.8}{1.2}{\meter}; if the human’s motion exceeds this range, the robot stops following. This range depends on the task and is fully configurable within the architecture. Crucially, it is not a limitation of the hardware or physical intelligence: if enlarged, ergonomic support is still guaranteed by the hardware design and the real-time adaptation of the physical intelligence. 
In Figure \ref{fig:collaborative}, we present the results of the human and robot collaborating to manipulate loads of different weights: empty, \SI{1}{\kilogram}, and \SI{2}{\kilogram}.
In Figure \ref{fig:collaborative} (a) we compare the human hands' height with the robot one. It can be noticed how the robot correctly follows the human in all three cases but that there are some delays in the robot's movements with respect to the human one, these are because there is no prediction of the human movements. Nevertheless, the robot consistently follows the human's movements, with an average error of \SI{0.0084}{\meter} and a mean maximum error of \SI{0.0339}{\meter} across the three experiments.
Figure \ref{fig:collaborative} (b) shows the maximum L5-S1 joint torque experienced by the human when manipulating loads of \SI{1}{\kilogram} and \SI{2}{\kilogram} and empty, with and without the robot's assistance. The stress on the lumbosacral joint is significantly reduced when the human collaborates with the robot. For instance, the peak torque decreases from $43.95 \text{ } Nm$  to $24.88 \text{ } Nm$ with the empty load, and a similar trend is observed for the \SI{1}{\kilogram} and \SI{2}{\kilogram} loads, where the peaks when collaborating with the robot drop from $50.66 \text{ } Nm$ to $ 25.77 \text{ } Nm$ and from  $44.88 \text{ } Nm$ to $31.82 \text{ } Nm$.
\subsubsection*{Locomotion tasks}
The optimization process that determined the primary shape of ergoCub also specifically aimed to enhance the robot locomotion abilities. Let us recall, in fact, that the optimization problem has a task specifically aimed at increasing the system bandwidth to improve the walking robustness. The synergy of the robot design with the locomotion architecture results in robust walking performances (in the sense of the ability to adapt footsteps for small pushes). Because of this, and compared to its predecessor iCub3, the ergoCub robot is able to walk faster and smoother; the maximal step length achieved is $\SI{0.35}{m}$ compared to $\SI{0.28}{m}$ on the iCub3 with a minimal step duration being $\SI{0.5}{s}$ compared to $\SI{0.8}{s}$ on the iCub3 using the same walking command and control architecture. The detailed comparison of the locomotion capability can be seen in Appendix C with the robots walking side by side given the same commands.
For optimizing the locomotion ability of the ergoCub robot, we employed the architecture instance represented in Figure \ref{fig:istances}b, disregarding the human model and focusing on the robot control architecture. 
The details of each component of the employed architecture, and the mathematics involved are deferred to Sections \ref{sec:method:locomotion} for the interested reader.
In the classical locomotion architectures, the trajectory adjustment layers uses \enquote{template} models (e.g. the Linear Inverted Pendulum LIPM, or more recently the centroidal dynamics model \cite{orin}) to compute the Center of Mass -    CoM -  and feet trajectories \cite{Giulio_benchmarking, Elobaid-RAL}, while the trajectory control layer provides individual position commands to the robot joints motors by solving an instantaneous inverse kinematics (or inverse dynamics) problem.
The trajectory generation and adjustment blocks run at   $\SI{50}{Hz}$, generating kinematically feasible nominal trajectories for the CoM and feet contact locations and timings respectively, while the trajectory control, being an instantaneous controller, running at $\SI{500}{Hz}$ solves a stack-of-tasks QP, generating desired joint velocities which are integrated to the low-level motor control boards. The solution is obtained using an off-the-shelf optimization solver, running on a 2.4 GHz Intel\textcopyright core i7 Ubuntu machine mounted inside the robot torso to serve as the main computational resource.
The robot locomotion capabilities in this context are tested in different scenarios, among which are the ones reported below
\begin{itemize}
    \item \emph{Locomotion with persistent disturbances:} in combination with the robot forearms and wrist hardware design, the detailed locomotion architecture allows for carrying the rated payload ($\SI{6}{kg}$) adjusting the nominal footsteps plan if necessary to counter the effect of this persistent disturbance.
    \item \emph{Locomotion with time-varying disturbances: } as can be seen in Figure \ref{fig:ergocubTasks}, and more precisely in multimedia video companion, the robot is subjected to increasing payload forces while a human is pushing on the    robot right upper arm (impulsive pushes estimated using an Newton-Euler method \cite{wbd_ref} propagating FT measurements to be  around $\SI{60}{N} - \SI{100}{N}$). The robot can complete the task (walking in a straight line), adjusting the steps within a nominal error bound if necessary to prevent falling (see Figure \ref{fig:ergocubTasks} - (a)).
\end{itemize}
\graphicspath{{Figures/}} 
\section{Discussion}
The work presented contributes to the state of the art by deriving a methodology for designing humanoid robots that explicitly account for human comfort and task requirements (e.g., walking stability) already at the hardware design stage, and not solely when implementing the robot behavior. In Section \ref{sec:results:ergocubRobot}, we presented the ergoCub humanoid robot, which is the outcome of the proposed modular architecture instances depicted in Figure \ref{fig:istances}a and \ref{fig:istances}b, which implements the paper methodology.
By integrating shared intelligence, coordinated action, and mutual awareness, with embodied cognition, namely the coupling between morphology and physical intelligence, the proposed methodology enables the joint design of robot morphology and control policies with respect to human-centered metrics. Within this perspective, the ergoCub robot represents a contribution in its own as a tangible realization of the methodology, demonstrating how human representations can be embedded into the robot’s body and physical intelligence to support ergonomic collaboration and robust locomotion.
The reported results illustrate the effects of this design approach on representative collaborative and locomotion tasks, providing evidence of the feasibility and practical relevance of the proposed framework, as well as the advantages of embedding human representation within the robot’s physical intelligence for human–robot collaboration. Indeed, the experimental campaign is designed to validate the feasibility and internal consistency of the proposed methodology as well as evaluate how the hardware optimization can be translated into a real humanoid robot. The focus is on demonstrating that embeeding task related objective during hardware optimization results in improved task performance. Particularly, for human-robot interaction, this translates into embedding human representations at the hardware optimization stage and results in improved ergonomy interaction. Broader user studies assessing long-term interaction fluency, subjective comfort across diverse populations, or clinical ergonomic validation represent complementary research directions that are not covered by the results presented in this paper. Additionally, the proposed optimization targets biomechanical risk metrics, which do not necessarily align with subjective perceptions of fatigue or comfort, further motivating the use of objective biomechanical indicators in the experimental campaign. The remainder of this section is organized as follows. In Section ~\ref{sec:discussion:human_rep}, we further elaborate on the adopted human representations, while in Section ~\ref{sec:discussion:human_prediction} we discuss possible extensions incorporating predictive components. Section ~\ref{sec:discussion:generalizability} addresses the generalizability of the hardware optimization framework and its potential extensions. In Section~\ref{sec:discussion:hardware_optimization}, we analyze how the hardware optimization was translated into a realized humanoid robot. Finally, Section ~\ref{sec:discussion:sim_to_real} proposes strategies to mitigate the sim-to-real gap arising from manufacturing process.
\subsection{Human representations}
\label{sec:discussion:human_rep}
The proposed modular architecture (Figure \ref{fig:istances}b) integrates both hardware optimization and robot physical intelligence to support locomotion and active human–robot collaboration through mutual partner awareness. This is achieved by explicitly incorporating human models into the hardware optimization process and by embedding models of human behavior within the robot’s control architecture. The human is represented as a multi-rigid-body system governed by a hierarchical control structure. This abstraction is embedded in both the hardware optimization and the robot’s physical intelligence. At the design stage, this level of abstraction enables the inclusion of human ergonomic metrics without overfitting to specific individual characteristics. The simplified morphology allows the optimization to capture general human-centered properties while remaining computationally tractable. At the control level, we adopt a unified framework in which both agents operate with internal representations of each other’s behavior: the robot relies on its human model, while the human relies on natural intuition and cognitive abilities to coordinate with the robot. Indeed, in collaborative scenarios, representing a partner’s actions and planning accordingly is essential for intelligibility and acceptability \cite{desantis2008}. Nevertheless, a fully detailed representation of human physical intelligence is unattainable due to the complexity of musculoskeletal and neuro-sensory processes. Simplifications are therefore necessary to enable real-time estimation from non-intrusive sensors and limited measurements. In our framework, the human physical intelligence is modeled using the same logical components as the robot, namely trajectory planning, adjustment, and control blocks. Although this symmetry is an abstraction, it captures the essential features required for ergonomic collaboration and enables seamless integration within the robot’s control architecture. As shown in Appendix A, the same modeling principles extend to more dynamic scenarios such as collaborative walking. Importantly, the results involve three distinct human subjects: one whose model informed hardware optimization, one performing collaborative payload lifting, and one engaged in collaborative walking. These subjects differ substantially in anthropometry and strength. Nevertheless, the ergonomic benefits are consistently preserved, demonstrating that the design-level optimization does not overfit a specific individual. Instead, inter-subject variability is accommodated through the robot’s physical intelligence, which integrates subject-specific modeling and real-time measurements. These choices are supported by both the present experimental validation and prior related work \cite{latella2019simultaneous, rapetti2020model, tirupachuri2021online, Latella2021, rapetti2023control}.
\subsection{Human representation improvements}
\label{sec:discussion:human_prediction}
We have analyzed in detail the rationale behind the adopted human representation and its role within the proposed robot physical intelligence. While this representation captures key structural and ergonomic aspects of human–robot collaboration, it currently models interaction in a purely reactive manner. Specifically, anticipatory behavior is not yet explicitly incorporated into the framework.
A conservative refinement could involve using the robot’s LCD interface to signal potentially unsafe postures, thereby encouraging proactive human correction without imposing explicit behavioral constraints; nevertheless, this would still rely on the real-time human measurement. A more comprehensive extension would integrate short-horizon human motion forecasting directly within the robot’s physical intelligence. Such forecasting could rely on data-driven models of human movement,for example, leveraging wearable sensing approaches as in \cite{guo2023online}, which are compatible with the sensing setup considered in this work.
Anticipatory behavior could be incorporated into the robot’s physical intelligence trajectory planning module. Short-horizon forecasts  would provide predicted future human states, from which corresponding ergonomic metrics could be estimated. The planning layer would then compute a robot trajectory that minimizes the expected ergonomic cost over the prediction horizon, effectively biasing behavior toward configurations that are most likely to reduce human musculoskeletal load in the near future. The resulting reference trajectory would be passed to the trajectory control layer, operating at a higher frequency to ensure accurate tracking while remaining responsive to real-time human measurements. Within this hierarchical structure, prediction informs planned behavior, while the robot trajectory control preserves robustness to modeling inaccuracies, sensing uncertainty, and unexpected human deviations by relying on real-time human measurements. Importantly, this extension affects only the physical intelligence layer and does not modify the hardware optimization stage, which embeds structural human representations at design time. The validity of the co-design methodology, therefore, remains independent of whether online interaction is purely reactive or augmented with predictive components.
\subsection{Generalizability across tasks and human partners}
\label{sec:discussion:generalizability}
In this work, we focused on payload lifting and walking tasks, which represent constrained yet relevant scenarios in industrial and healthcare contexts. This choice enabled a clear analysis of how the proposed methodology translates from an abstract optimization framework into a realizable robotic platform, and allowed us to evaluate its effectiveness on concrete tasks. Nevertheless, the methodology can be extended to accommodate different task requirements by encoding them through appropriate objective functions and constraints. It can also incorporate additional hardware and control related goals, such as power consumption, mass distribution, end-effector workspace specifications, and dynamic behaviors (e.g., running). Importantly, human-related ergonomic metrics and locomotion-related objectives were jointly included within a single optimization problem, already demonstrating the flexibility of the framework to address multiple task requirements simultaneously. A promising extension would be to encode collaborative scenarios involving multiple human partners and heterogeneous payloads. This could be achieved by extending the same optimization problem, specifically by replicating the coupled human–robot–load dynamics for each subject or load of interest. Since the optimization is performed offline to identify optimal hardware parameters, the resulting computational overhead would likely remain tractable. However, real-time human measurements and subject-specific models embedded in the robot’s physical intelligence would still be required to ensure smooth interaction and preserve the generalizability of the proposed approach. More generally, although the experimental validation focuses on structured collaborative tasks, the architecture itself is not restricted to predefined motion patterns. During interaction, the robot continuously estimates human kinematics and dynamics through real-time sensing and updates its internal human model accordingly. This enables the robot’s physical intelligence to react to variations in human motion and behavior during collaboration. Future extensions incorporating predictive models of human motion, as discussed in Section \ref{sec:discussion:human_prediction}, could further enhance the system’s ability to anticipate evolving human strategies in more dynamic and less structured collaboration scenarios.
\subsection{From optimized models to a real robot hardware}
\label{sec:discussion:hardware_optimization}
The distinctiveness of the proposed hardware optimization lies in its explicit use of human–robot coupled dynamics and non-linear optimization, directly linking hardware characteristics to task performance. By incorporating ergonomic metrics, the human factor is addressed at the design stage rather than only during behavior implementation. A limitation of this approach is the sensitivity of non-linear methods to initial conditions and local minima, which we mitigated by initializing with iCub3, thereby starting from a proven design, ensuring manufacturability. To keep the problem tractable, we considered multiple static task configurations, leaving motion-related ergonomics to the physical intelligence component. Simplifications in modeling, however, led to some mismatches between the optimized design and the manufactured ergoCub. As shown in Figure \ref{fig:hardware_optimization_output} (e), ergoCub differs from the optimization output in height, weight, and slightly in limb lengths. These discrepancies arise from several factors for instance, the head is larger than in the optimized model, since an LCD screen was added to display emotions, increasing the overall height compared to the original iCub design. Regarding the robot weight, the optimization assumed the densities constant along the link, whereas during manufacturing, we reused iCub3 electronics and motors \cite{dafarra2024icub3}
and employed metal extenders to lengthen the limbs. This difference, combined with the manufacturing optimization process, resulted in lighter components and produced a robot approximately \SI{0.25}{\meter} taller than iCub3, yet weighing only 56.70 kg, i.e., about 4 kg more than iCub3. Compared to the optimization output, which predicted a mass of 70 kg under simplified density assumptions, the final robot is substantially lighter thanks to a dedicated manufacturing optimization process. The legs and torso of ergoCub are \SI{0.01}{\meter} shorter than the optimized design because the optimization simplified links into basic shapes and omitted assembly details such as covers, electronics, and wiring. The accumulated small mismatches in building the robot led to slight differences in proportions during construction.
The simplifications introduced were necessary for tractability but do not undermine the generality of the method. In practice, assumptions such as simplified shapes or uniform densities are relaxed during manufacturing, where each link is refined at the sub-component level while preserving the whole-body properties identified by optimization. This enables attainment of the global link optimum, even if equivalent masses or volumes are achieved through adjustments in the smaller link components.
Despite its challenges, the approach yielded tangible results, with the ergoCub robot successfully retaining the ergonomic enhancements from the optimization process.
\subsection{Bridging the sim-to-real gap}
\label{sec:discussion:sim_to_real}
Discrepancies between the optimization output and the manufactured robot primarily arise from three sources: (i) geometric simplifications adopted to ensure tractable nonlinear optimization, (ii) simplified inertial assumptions, such as uniform density distributions, and (iii) practical manufacturing constraints that prevent the physical hardware from exactly matching the optimized design (e.g., component reuse, wiring, protective covers, and mechanical tolerances). Several strategies could reduce the sim-to-real gap. Regarding geometric simplifications, the hardware optimization model could be refined by decomposing links into subcomponents (e.g., motors, electronics, structural elements, and covers), although this would increase problem complexity. Concerning inertial assumptions, material-dependent density distributions could be incorporated into the optimization and treated as design variables. While this would remain an approximation, since link density is not perfectly uniform, it would yield a more accurate representation of mass distribution and potentially improve optimization fidelity. With respect to manufacturing constraints, the sim-to-real gap can be further reduced through progressive model refinement. An initial optimization stage may rely on simplified link representations to identify globally consistent proportions, mass distributions, and dynamic properties. As the mechanical design evolves, subsequent optimization stages can incorporate increasingly detailed geometric and inertial descriptions, with the optimzation that follows the stages of the mechanical design. Another approach could involve calibration loops introduced after prototype realization: experimentally measured sub-component parameters, such as inertial parameters and joint-level actuation characteristics, can be fed back into the optimization model, enabling dynamic parameter re-identification and refinement of task-level objectives. Surrogate modeling techniques can also be used to improve physical accuracy while reducing computational burden. Nevertheless since hardware parameters constitute the decision variables of the nonlinear optimization, any surrogate model must be formulated explicitly as a function of those same variables. This ensures that detailed hardware characterization can be introduced without removing hardware parameters from the optimization search space. 
\newpage
\section*{Declarations}
\paragraph{Data Availability:} The data supporting the findings of this study are publicly available in the paper repository at \url{https://github.com/ami-iit/paper_sartore_2025_ergocub_nature_machine_intelligence}~\cite{sartore_ergocub_nature_release_2025}. For further inquiries regarding the data, please contact the corresponding authors at \href{mailto:carlotta.sartore@gbionics.ai}{carlotta.sartore@gbionics.ai} and \href{mailto:daniele.pucci@gbionics.ai}{daniele.pucci@gbionics.ai}.
\paragraph{Code Availability:}  The code developed to derive the results of the presented work is available at \url{https://github.com/ami-iit/paper_sartore_2025_ergocub_nature_machine_intelligence} \cite{sartore_ergocub_nature_release_2025}.
\paragraph{Acknowledgements:} The authors acknowledge Mattia Salvi, Davide Gorbani, Camilla Gallina and Dario Maria Sortino for their technical support during the experimental phases of this work, and Francesca Bruzzone and Bruno Trombetta for their support in video production and material preparation.
\paragraph{Author contributions Statement:}
\begin{itemize}
    \item \textbf{Conceptualization}: C.S., L.R., M.E., L.F., F.D., G.T., L.N, M.M,  D.P.
    \item \textbf{Funding acquisition}: M.M., D.P.
    \item \textbf{Investigation}: C.S., L.R., M.E.,G.R., S.D.,N.A.P., I.S., P.M.V., S.T., L.F.
    \item \textbf{Methodology}: C.S., L.R., M.E., D.P.
    \item \textbf{Project administration}: D.P.
    \item \textbf{Software}: C.S., L.R.,M.E., G.R., S.D., S.T., U.P., 
    \item \textbf{Supervision}: L.R., L.N., M.M., D.P.
    \item \textbf{Validation}: C.S., L.R., M.E., G.R., S.D., I.S., P.M.V.
    \item \textbf{Visualization}: C.S., L.R., M.E., N.A.P.
    \item \textbf{Writing}: C.S., L.R., M.E., D.P. 
\end{itemize}
\paragraph{Competing interests Statement:} The authors declare that they have no competing interests.
\paragraph{Funding:} The paper was supported by the Italian National Institute for Insurance against Accidents at Work (INAIL) ergoCub Project.
\newpage
\clearpage

\renewcommand{\thefigure}{\arabic{figure}}
\begin{figure}[!tbp] 
  \centering

    \includegraphics[trim={0.0cm 5cm 0cm 0cm},clip,width=\linewidth]{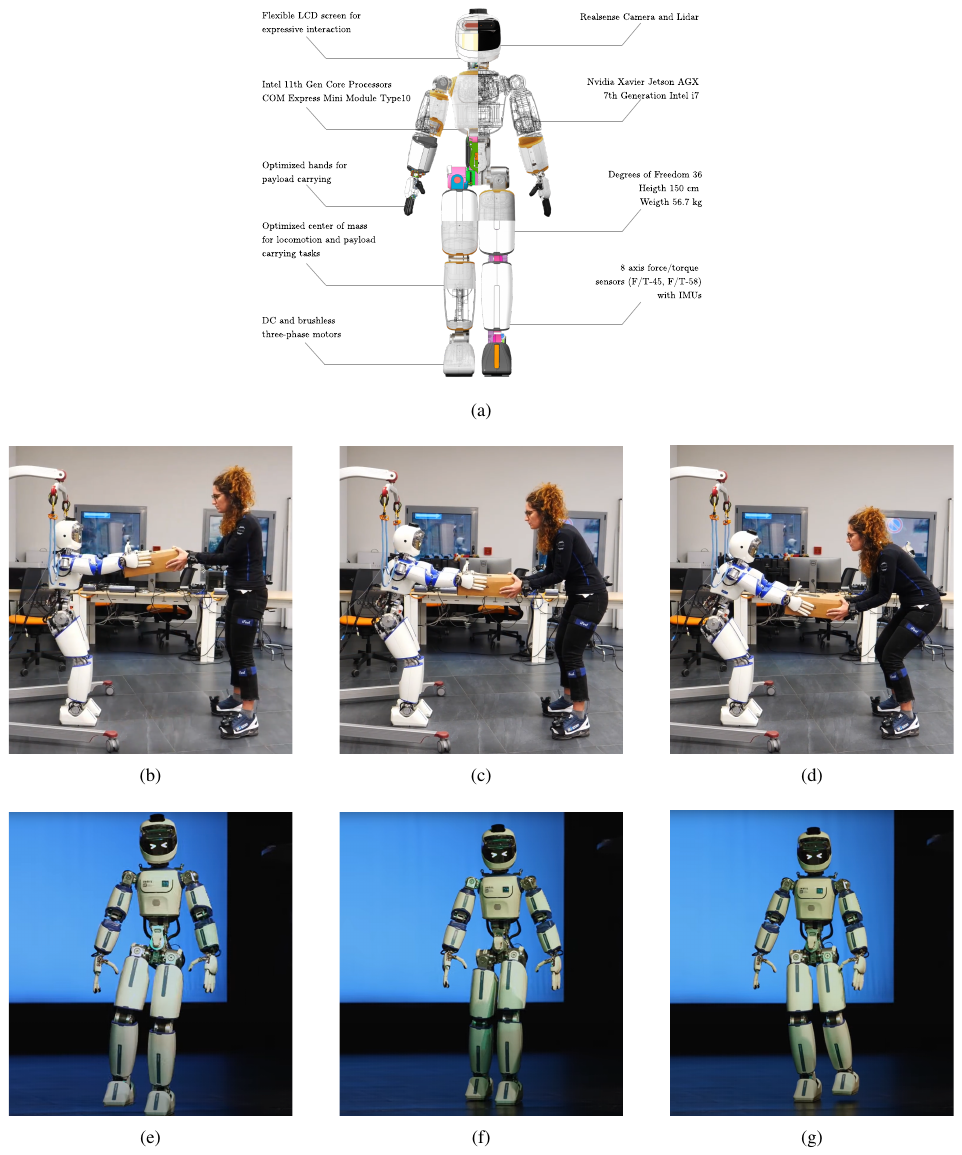}  

  \caption{The ergoCub robot, a humanoid robot developed to minimize workers' fatigue and risks in collaborative tasks for industry and healthcare, and endowed with a degree of shared embodied intelligence. In (a), the main characteristics of the ergoCub robot. In (b)–(d), The ergoCub robot collaborates with a human to perform a payload lifting task, adapting its motion to the human partner and monitoring the human’s ergonomic state to reduce physical stress.  In (e)–(g), the ergoCub robot walks during a live demo on stage.}
  \label{fig:ergocub}
\end{figure}

\newpage
\begin{figure}
     \centering
    \includegraphics[width=1\textwidth,trim={0.0cm 0.0cm 0.0cm 0cm},clip]{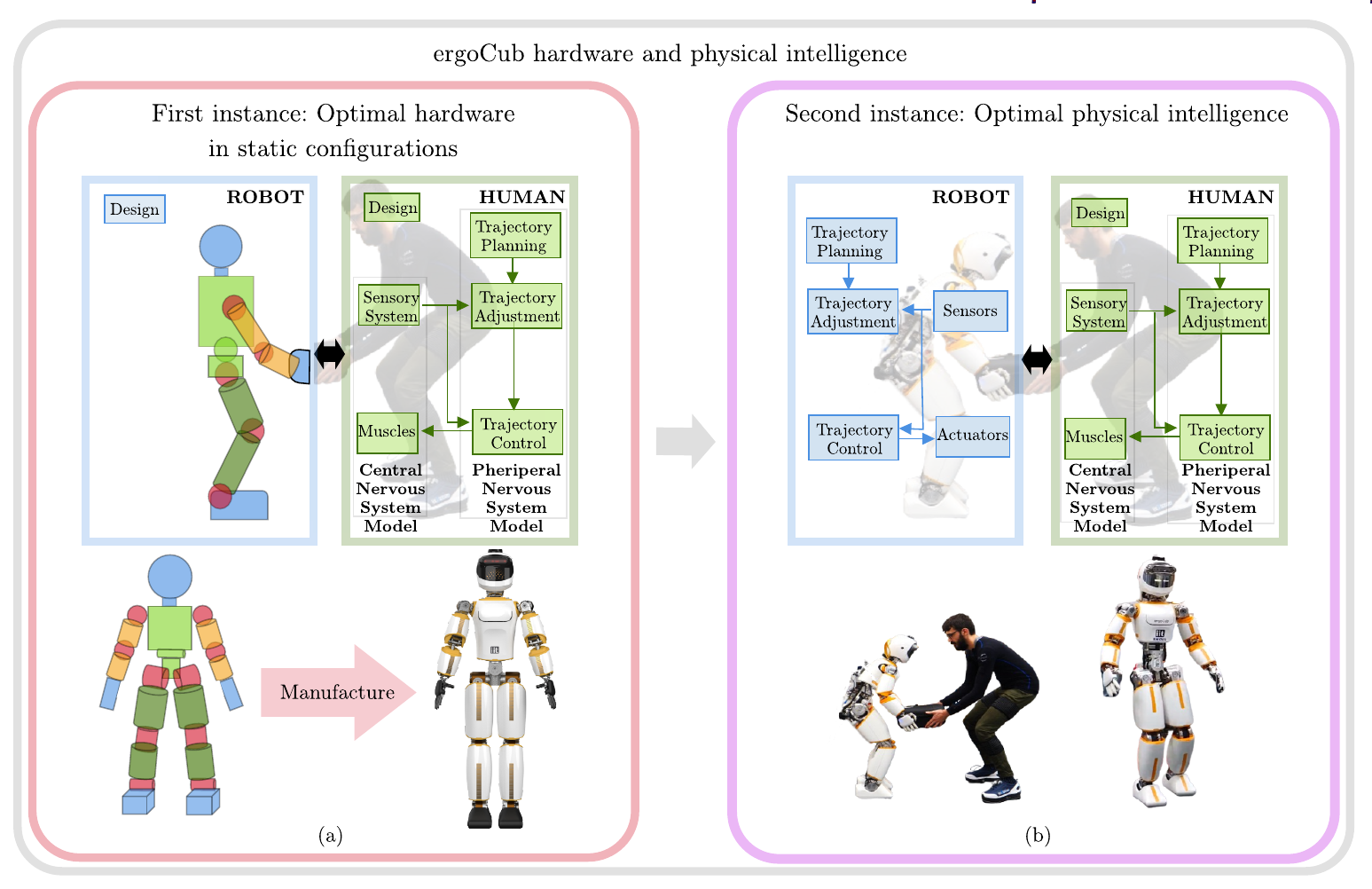}
    \caption{The shared embodied intelligence architecture instances to optimize robot hardware and physical intelligence leveraging human-robot interaction models that are parametrized with respect to robot hardware parameters. Within this architecture, both human and robot physical intelligence are structured within a hierarchical control framework, establishing symmetry between the two agents. Figure (a) illustrates the architecture instance that identifies optimal robot hardware parameters for static collaborations with human and walking tasks. The output consists of optimal robot link lengths to enhance human ergonomics and robot walking capabilities. Such output has been used to manufacture the real ergoCub robot. In Figure (b), the architecture instance to optimize the ergoCub physical intelligence given the optimal hardware. The output enables the ergoCub robot to collaborate with humans and walk robustly.}
\label{fig:istances}
\end{figure}
\newpage
\begin{figure*}[!t]
\centering
\includegraphics[trim={0.0cm 8cm 0cm 0cm},clip,width=\linewidth]{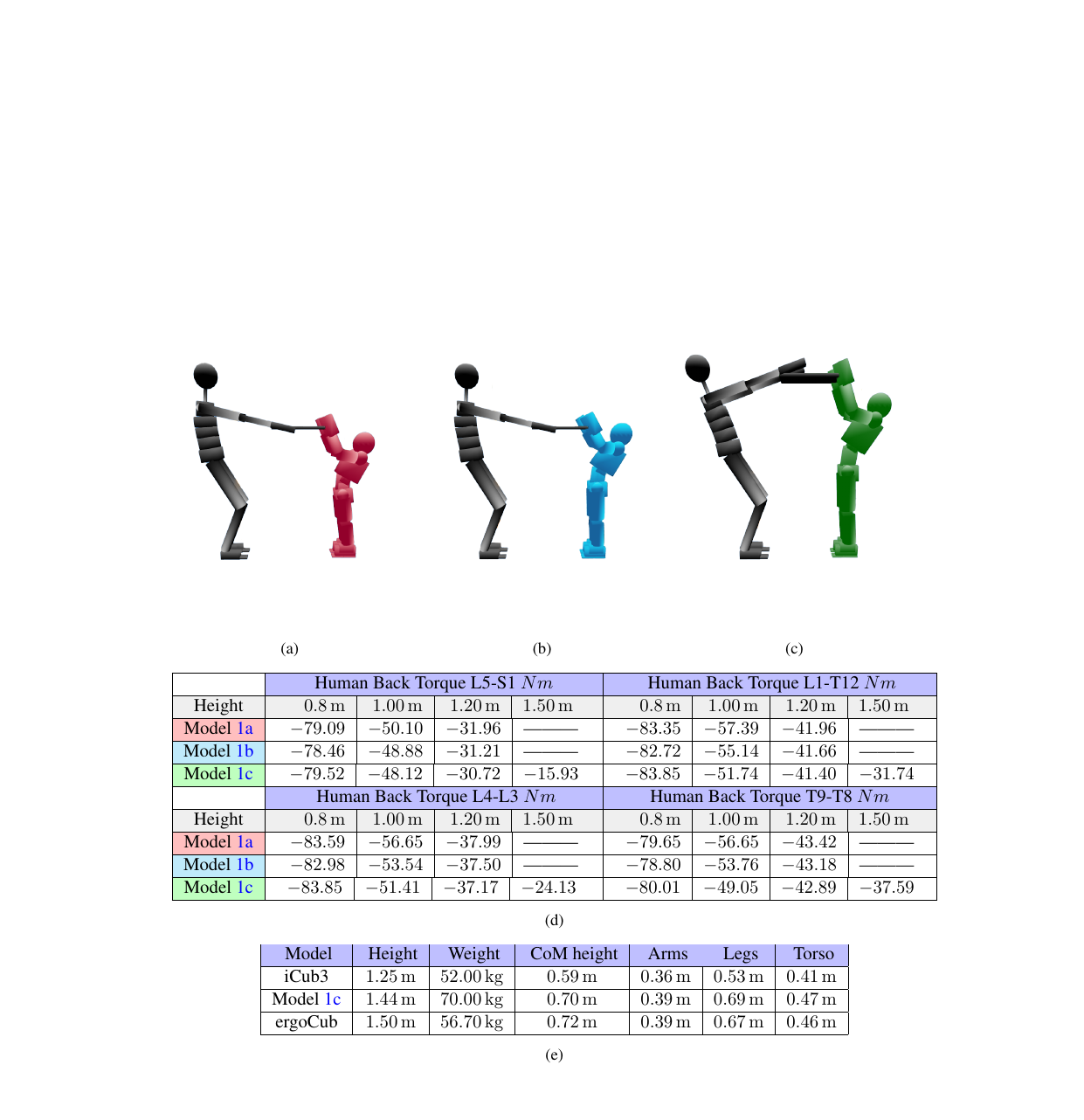}  
\caption{Comparison between iCub3,  different humanoid robot models output of the presented optimization framework and ergoCub.  In (a), the iCub3 simplified robot collaborates with a human. In (b) and (c), we present the humanoid robot outputs optimized for collaborative payload lifting, with load height ranges of \SIrange{0.8}{1.2}{\meter} and \SIrange{0.8}{1.5}{\meter}, respectively. In (d), we illustrate the human back torques when holding a box at different heights using the models depicted in Figures (a), (b), and (c) in green we highlight the model used to define the ergoCub robot link lengths. In (e) the main dimensions of ergoCub, iCub3, and the optimization output for Figure (c), used to define the ergoCub limb lengths. }
\label{fig:hardware_optimization_output}
\end{figure*}
\newpage

\begin{figure}  
     
\includegraphics[trim={0.0cm 8cm 0cm 0cm},clip,width=\linewidth]{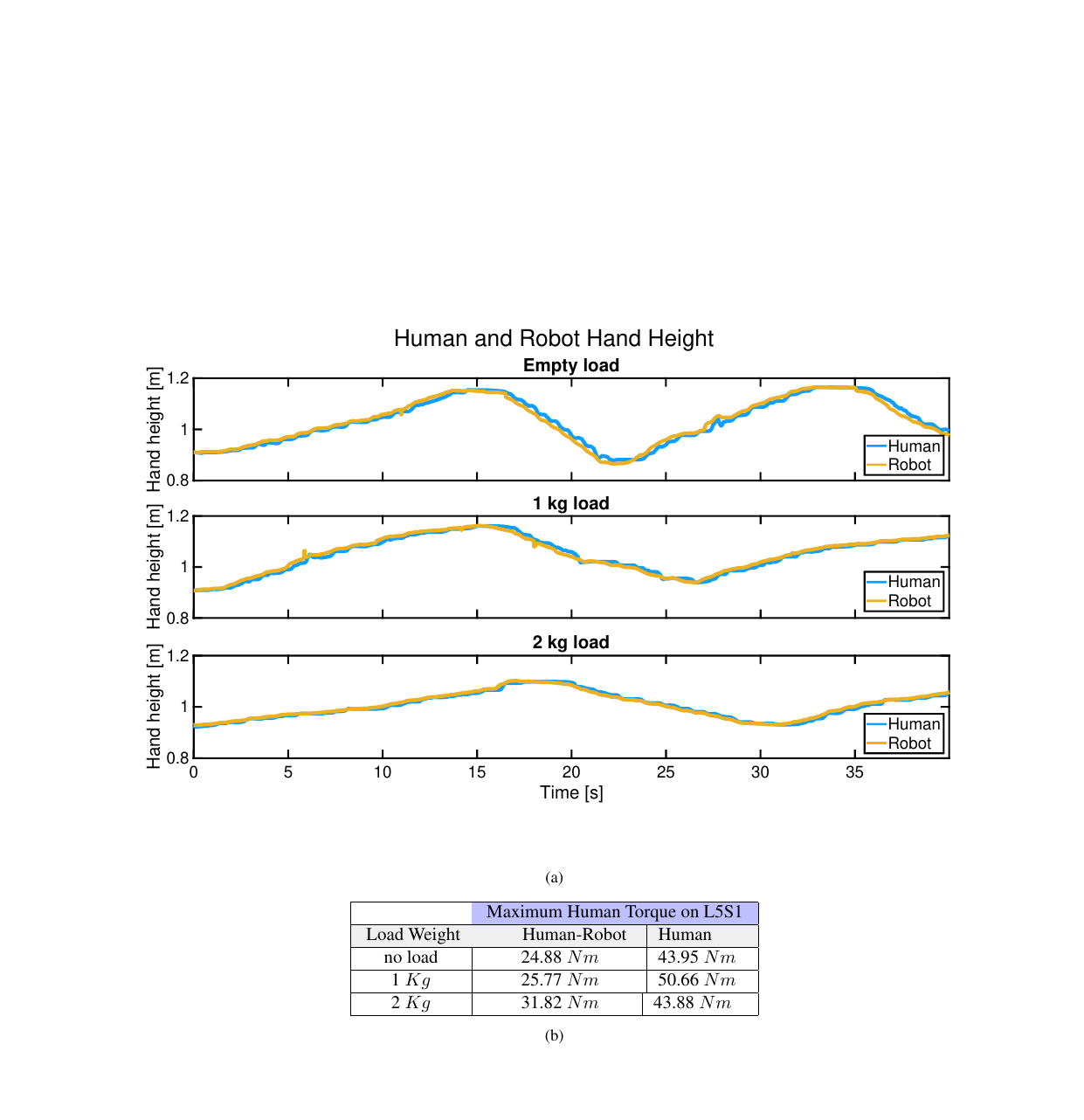}  
 \caption{ 
 Comparison of hand trajectories and lumbosacral joint torques during human-robot collaborative payload lifting with ergoCub robot. In (a) Comparison of hand height trajectories between the human and the robot during the lifting task with three different payloads. In (b), the maximum estimated torque on the human's lumbosacral joint (L5-S1) during the payload lifting task with three different weights, comparing collaboration with the robot versus performing the task alone.
 }
    \label{fig:collaborative}
\end{figure}
\newpage
\begin{figure}[!htbp]
\includegraphics[trim={0.0cm 15cm 0cm 0cm},clip,width=\linewidth]{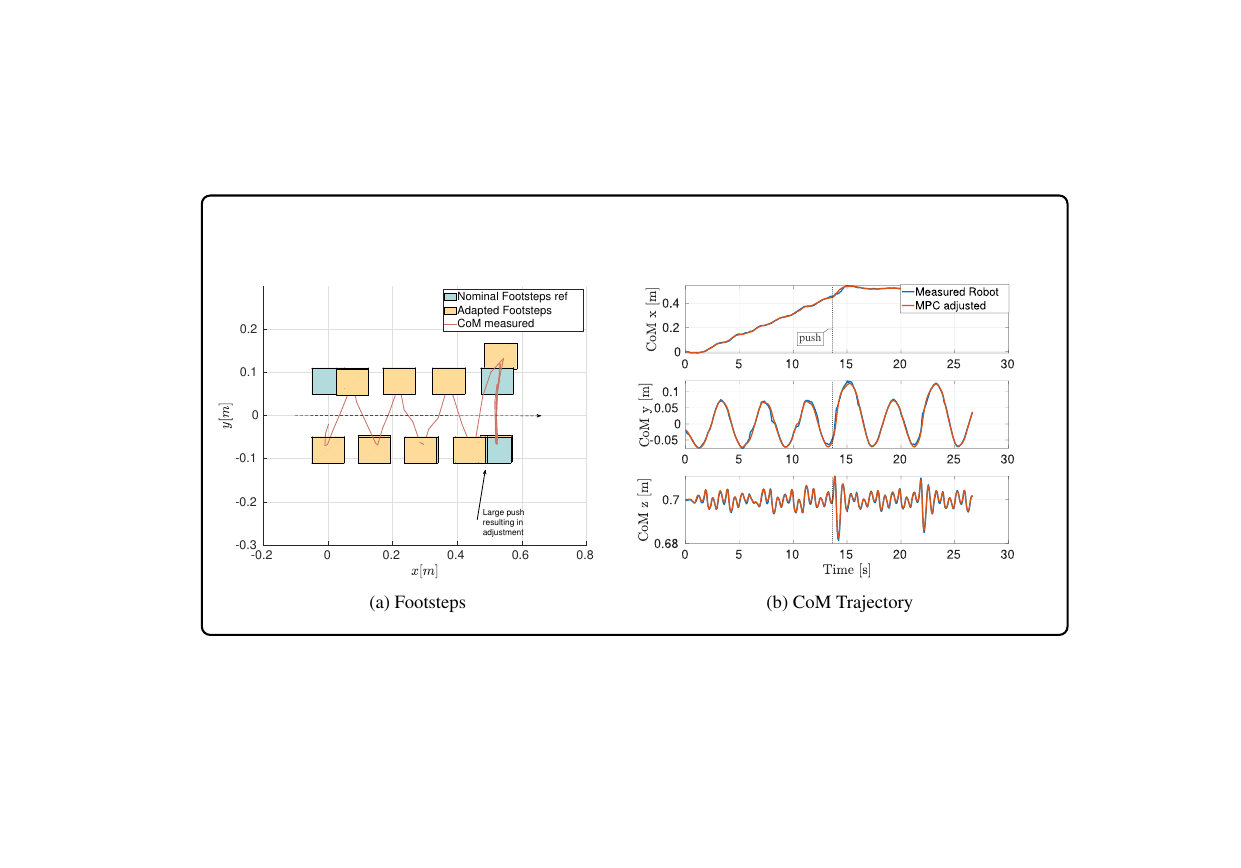}  
\caption{Figure showing the tracking performances of the ergoCub Controller while executing a walking-in-straight-line task subject to pushes and carrying a heavy box. $(a)$ the reference and actual footsteps. The adaptation at the end of the sequence is due to push exerted by the human, with direction of walking given by the dashed arrow, $(b)$; the CoM trajectory tracking performance, with the big push corresponding to a larger $y$ component adjustment.}
\label{fig:ergocubTasks}
\end{figure}
\newpage
 \begin{figure}
        \centering
        \includegraphics[width=0.80\textwidth]{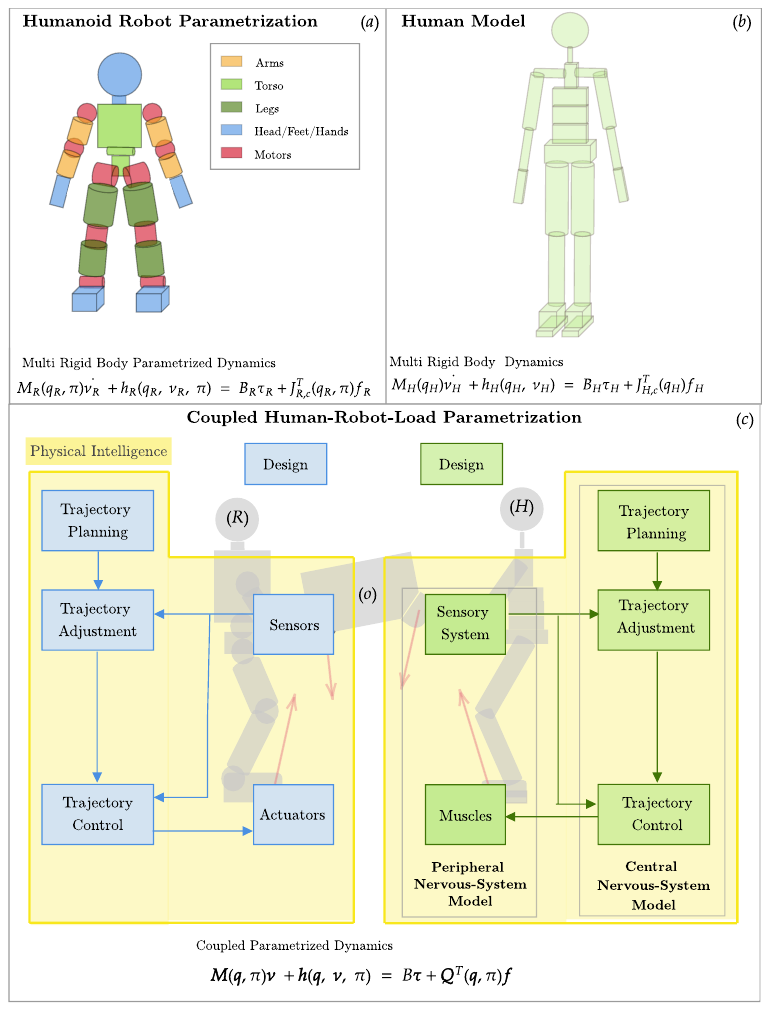}
        \caption{ Shared embodied intelligence framework. In (a), we depict the parametrization of humanoid robot dynamics based on hardware parameters. (b) Shows the human modeled as a multi-rigid body system. (c) Illustrates the coupled dynamics of human-robot-load, parametrized with respect to robot hardware parameters. In (c), we also display the physical intelligence architecture employed, consisting of trajectory blocks providing joint-level inputs to muscles/actuators using sensor/sensory system feedback. For the human, physical intelligence is divided into the peripheral and central nervous systems.}
        \label{fig:diagram_robot_parametrization}
    \end{figure}
\newpage
\newpage
\clearpage
\graphicspath{{MaterialsAndMethods/Figures/}} 
\section{Methods}
At the methodological level, our goal is to propose a shared embodied intelligence architecture to enable effective physical collaboration between robots and surrounding humans, namely, identify robot hardware and control parameters to optimize quantities of the human collaborator. 
A high-level sketch of the framework is depicted in Figure~\ref{fig:diagram_robot_parametrization}. We design the architecture to be modular, being composed of different blocks suitable for various robot tasks performed with or without a human collaborator.    The first key element is the \emph{agent design}, hence human body properties and robot hardware characteristics. The second key element of the architecture is representations of both the human and the robot's physical intelligence, thus the blocks that define the agent motion to achieve the desired behavior.
   For both the human and the robot,  we employed a hierarchical control architecture to model the physical intelligence,  establishing a symmetry between the two agents. 
The key blocks constituting both human and robot physical intelligence include:
(i) \emph{muscles/actuators}, which model joint-level actuation in both the robot and the human;
(ii) \emph{sensors/sensory systems}, which perceive information from the environment and the agent’s state; and
(iii) \emph{trajectory planning}, \emph{trajectory adjustment}, and \emph{trajectory control} blocks, which respectively generate, refine, and execute trajectories to achieve the desired behavior.
For what concerns the human,  for the sake of simplicity, we associate the peripheral nervous system with the joint control loop and the central nervous system with whole-body control.
Building upon this modular modeling, we developed a parametrized architecture incorporating environmental factors such as the interaction between different agents or the agents and the environment. We present two instances of this architecture, as illustrated in Figure \ref{fig:istances}, whose outputs culminate in the realization of the ergoCub robot, as presented in the previous Section \ref{sec:Results}. The implemented architecture instances are obtained by activating specific sub-modeling components, and each of these instances targets specific objectives and differs in terms of optimized parameters. The ultimate objective of each instance is to identify an architecture configuration that realizes the envisioned shared embodied intelligence.  
In the following sections, we delve into the different sub-blocks that compose the architecture and the instances defined. 
The rest of the sections are organized as follows: in Section \ref{sec:material:modelling}, we detail the modeling employed to represent the human, the robot, and the human-robot interaction, as well as the models employed in the robot physical intelligence and the hardware optimization processes. Subsequently, in Section \ref{sec:methods:morphology_optimization}, we delve into the optimization process that crafted the ergoCub robot, as presented in Section \ref{sec:results:ergocubRobot}. Moving on to Section \ref{sec:methods:shared_control_framework}, we explore the architecture instances that define the ergoCub physical intelligence, which confers upon the ergoCub robot the capability to engage in physical collaboration with humans and exhibit walking abilities, as shown in Section \ref{sec:results:ergoCub_physical_intelligence}.
\subsection{Modeling}
\label{sec:material:modelling}
In the context of this work, we define physical collaboration as the ability of humanoid agents, whether they be humans or humanoid robots, to interact with the environment while jointly lifting objects. To achieve the shared embodied intelligence envisioned in this study, it is essential to examine the mutual influence each agent exerts on the other, as well as the impact of hardware and control parameters on this interaction.
In these sections, we provide a detailed description of the humanoid robot dynamic parametrization with respect to robot hardware parameters, together with the exploited human model. Furthermore, we formulate the coupled dynamics, building upon the individual representations of both the robot and the human, that describe their reciprocal interaction and the interaction with the environment.
These models are leveraged to identify the optimal hardware design and physical intelligence required to attain shared embodied intelligence. Before delving into the details properly, we define the following basic notations and nomenclature which will be used throughout the rest of this section.
\begin{itemize}
    \item For an element of a normed Euclidean 3D space, $x \in \mathbb{R}^3$, $x^\wedge = S(x) : \mathbb R^3 \to \mathfrak{so(3)}$ returns the skew symmetric matrix form of $x$.
    \item Given an inertial frame, $\mathcal{I}$, and a frame attached to the rigid body (base frame)   $\mathcal{B}$, $\prescript{\mathcal{I}}{}{p}_{\mathcal{B}} \in \mathbb{R}^3$ and $\prescript{\mathcal{I}}{}{R}_{\mathcal{B}} \in SO(3)$ denote the position and the orientation of the \textit{base frame} respectively.
    \item Whenever the superscripts are dropped, quantities are referred to the inertial frame, e.g. $p_{CoM}$ denotes the position of the CoM of the model referred to the inertial frame.
    \item Denote by $\mathbb{I}_3, \ 0_3$ the identity and zero matrices of dimension 3 respectively.
\end{itemize}
\subsubsection{Humanoid robot parametrized dynamics}\label{robot-parametrized_dynamics-section}
To parameterize the robot dynamics with respect to the robot hardware parameters, we establish the relationship between the hardware parameters, namely link lengths, and densities, and the inertial characteristics of the rigid body. For any given rigid body, its inertial properties are described by the density $\rho \in \mathbb{R}^+$, and $\mathcal{V} = \mathcal{V}(l) \subset \mathbb{R}^3$, representing a region in the 3D Euclidean space where $\rho \neq 0$ and dependent on the body shape and a set of dimensional parameters $l \in \mathbb{R}^{n_l}$. This parametrization allows us to express the rigid body mass $m$, the inertia $I$ and the rigid body center of mass $c$ computed with respect to the rigid body frame $\mathcal{L}$ as functions of $\rho$ and $l$, as per Equation \eqref{eq:rigid-body-parametrization}, where $r$ is expressed with respect to the rigid body frame $\mathcal{L}$
\begin{IEEEeqnarray}{RLC}
\IEEEyesnumber
\label{eq:rigid-body-parametrization}
&m(l, \rho) = \iiint_{\mathcal{V}(l)}\rho(r)\cdot dr, \nonumber\\
&I(l, \rho) = - \iiint_{\mathcal{V}(l)} \rho(r)\left[ S(r)\right ]^2\cdot dr,\IEEEeqnarraynumspace \\[6pt] 
&c (l, \rho) = \frac{\iiint_{\mathcal{V}(l)}r\rho(r) \cdot dr}{m(l,\rho)}. \nonumber
\nonumber
\end{IEEEeqnarray}
We consider basic shapes such as spheres, cylinders, and boxes. We parametrize their geometry using $l_m \in \mathbb{R}^+$ as a multiplier that elongates the shape along its principal directions and we assumed $\rho$ to be uniform across all body points. 
With this established relationship, we then parameterize the rigid body dynamics with respect to $l_m$ and $\rho$ as per Equation \eqref{eq:rigidBodyDynamicsWithHardw}, following the notation of \cite{traversaro2017thesis}:
\begin{equation}
\label{eq:rigidBodyDynamicsWithHardw}
M(l_m, \rho)a^g +\mathrm{v}\bar{\times}^*M(l_m, \rho)\mathrm{v} = f,
\end{equation}
 \noindent where 
 \begin{itemize}
    \item $\mathrm{v} = \left[ \prescript{\mathcal{I}}{}{\dot{p}}_{\mathcal{L}}, \prescript{\mathcal{I}}{}{\omega}_{\mathcal{L}} \right] \in \mathbb{R}^6$ denotes the linear and angular velocity of the body.  
     \item $a^g \in \mathbb{R}^{6}$ is the \emph{proper} body acceleration thus the difference between the body acceleration and the gravity acceleration, i.e.: 
    $a^g = \dot{\mathrm{v}} - \begin{bmatrix} \prescript{\mathcal{I}}{}
    {R}_{\mathcal{L}}^Tg \\ 0_{3} \end{bmatrix}$, where $\prescript{\mathcal{I}}{}
    {R}_{\mathcal{L}}\in SO(3)$ is the rotation matrix of the frame $\mathcal{L}$ with respect to the inertial reference frame $\mathcal{I}$ 
    \item $M \in \mathbb{R}^{6\times6}$ is the 6D inertia matrix defined as: 
 \begin{equation}
 \label{eq:rigidBodyMass}
     M(l_m, \rho) = \begin{bmatrix}m\mathbb{I}_3 & -mS(c) \\mS(c) & I\end{bmatrix},
 \end{equation}
where $m$ and $I$ are computed as per Equation \ref{eq:rigid-body-parametrization}.
\item $\mathrm{v}\bar{\times}^*$ is the 6D force cross product operator \cite{featherstone2014rigid} defined as: 
        $\mathrm{v}\bar{\times}^*=\begin{bmatrix}
            S(\omega) & 0 \\
            S(v) & S(\omega)
        \end{bmatrix}$
 \end{itemize}
The same approach as described in Equation \eqref{eq:rigidBodyDynamicsWithHardw} was applied to each link of the robot kinematic chain. This enables us to define ${\pi}$ as the set of hardware parameters associated with the robot links, encompassing the length multiplier $l_m$ and density $\rho$ for each link.
To formulate the robot's dynamics, we apply the Euler-Poincarè formalism \cite{Marsden2010} to the kinematic chain. This results in a comprehensive set of differential equations, accompanied by holonomic constraints that characterize the robot dynamical behavior parametrized with respect to the set of hardware parameters $\pi$:
\begin{IEEEeqnarray}{RLC}
\IEEEyesnumber
\label{eq:constrained-dynamic-hardware-param}
& M(q, \pi) \dot{\nu} + h(q,\nu,\pi) = B {\tau} + J_c^T(q,\pi) f,  \label{eq:constrained-dynamic-hardware-param:dynamics} \IEEEyessubnumber \\
& J_c(q,\pi) \nu = 0,\IEEEyessubnumber  \label{eq:constrained-dynamic-hardware-param:constraints} 
\yesnumber
\end{IEEEeqnarray}
where: 
\begin{itemize}
\item $n_R$ is the number of the robot \emph{joints} with one degree of freedom that connect $n_R+1$ rigid bodies, i.e. \emph{links} that composes the robot.
\item $q = \left[\prescript{\mathcal{I}}{}{p}_{\mathcal{B}}, \prescript{\mathcal{I}}{}{R}_{\mathcal{B}}, s\right ]$ represents the system configuration, where $s \in \mathbb{R}^{n_R}$ is the joints configuration and $\mathcal{I}$ is the reference inertial frame. 
\item $\nu \in \mathbb{V}$  is the model velocity and is defined as $\nu = \left[\prescript{\mathcal{I}}{}{\mathrm{v}}_{\mathcal{B}}, \dot{s}\right]$ where $\prescript{\mathcal{I}}{}{\mathrm{v}}_{\mathcal{B}}=\left[\prescript{\mathcal{I}}{}{\dot{p}}_{\mathcal{B}}, \prescript{\mathcal{I}}{}{\omega}_{\mathcal{B}}\right] \in \mathbb{R}^6$ denotes the linear and angular velocity of the \textit{base frame}, and $\dot{s}$ denotes the joint velocities.
\item $M \in \mathbb{R}^{\left(n_R+6\right) \times \left(n_R+6\right)}$ is the mass matrix, 
\item $h \in \mathbb{R}^{n_R+6}$ is the term accounting for Coriolis and gravity forces.
\item $B = \left[0_{n_R \times 6},\mathbb{I}_n\right]^T$ is a selector matrix.
\item ${\tau}  \in \mathbb{R}^{n_R}$ is a vector representing the robot's joint torques.
\item $f \in \mathbb{R}^{6n_c}$ represents a vector containing $n_c$ wrenches stacked.
\item $J_c \in \mathbb{R}^{n_R+6 \times 6n_c}$ is the Jacobian of the contact frames.
\end{itemize}
In this work, we consider a robot composed solely of basic shapes — spheres, cylinders, and boxes. Therefore, starting with the iCub3 robot topology, we designed the model depicted in Figure \ref{fig:diagram_robot_parametrization}a.
\subsubsection{Human modeling}
\label{sec:methods:human-modelling}
   A comprehensive description of the human body and its physical intelligence is nearly impossible, even with vast amounts of data. Therefore, it is crucial to select models that effectively capture key aspects of the human body and behavior based on specific objectives.
In this work, we utilize models of the musculoskeletal system, which governs body structure and movement, and the nervous system, which processes sensory information and generates motor commands \cite{ingram2011}. 
   \noindent \textbf{Musculoskeletal system}
   \noindent We begin by modeling the human skeleton as a floating-base multi-body system with $N_H$ links connected through $n_H$ joints. In the context of this work, $n_H =34$ joints, 6 per arm, 9 in each leg, and 4 in the back. Each link corresponds to a segment of the human skeleton represented as a basic geometry, that prioritizes fast processing over fidelity, based on the model proposed in \cite{latella2019simultaneous}. The mass is distributed on different body districts based on anthropometric tables \cite{hanavan1964mathematical, Herman2016}. The resulting model is depicted in Figure \ref{fig:diagram_robot_parametrization}b. The skeletal muscles act on the human skeleton by contracting and consequently transmitting the force through the tendons to the bones. Considering that there are more than 600 skeletal muscles \cite{garzaulloa2018_2}, in the proposed model, the effect of muscle groups is synthesized by the resulting torques $\tau_H \in \mathbb{R}^{n_H}$ applied to the system joints. 
 The human dynamics is therefore described with the same set of equations presented in Equation \ref{eq:constrained-dynamic-hardware-param} for the robot dynamics, but disregarding the dependency on the hardware parameters $\pi$. The human torques are computed online with a  Maximum a Posteriori (MaP) estimation as per \cite{latella2019simultaneous}, exploiting the contact wrench measurements given by the FT on the shoes and the output of the Inverse kinematic estimation that computes the joint configuration given the iFeel and tracker sensors output as per \cite{rapetti2020model}. More in detail, the system dynamics is rearranged  into a compact matrix form of Equation  \eqref{eq:MAPEST:estimation}
\begin{equation}
       \label{eq:MAPEST:estimation}
    \begin{bmatrix}
        Y(s_H) \\
        D(s_H)
    \end{bmatrix} d + \begin{bmatrix}
        b_Y(s_H, \dot{s}_H) \\
        b_D(s_H, \dot{s}_H) 
    \end{bmatrix} = \begin{bmatrix}
        \gamma\\
        0
    \end{bmatrix},
\end{equation}
    \noindent where $Y$ and $D$ are matrices accounting for the sensor measurements and the system dynamics respectively, $b_Y$ and $b_D$ are the associated bias term. $\gamma$ is a vector collecting the sensor measurements, in our case, it contains IMU and Force Torque (FT) measurements from the iFeel sensors. The vector $d$ includes the dynamic and kinematic quantities of the links and joints such as the proper sensor acceleration, the internal joint wrench, and the joint acceleration. The algorithm maximizes the probability of $d$ given the sensor measurements. which leads to : 
\begin{equation}
       \begin{bmatrix}
        \mu_{d|\gamma} &\Sigma_{d|\gamma} 
    \end{bmatrix} =\underset{d}{\text{argmax}} \left(\mathcal{P}\left(d|\gamma\right)\right),
\end{equation}
   \noindent where $\mu$ and $\Sigma$ are respectively the mean and the covariance matrix, while $\mathcal{P}\left(d|\gamma\right)$ is the probability of $d$ conditioned on $\gamma$. 
   \noindent\textbf{Nervous system}
\noindent
   The nervous system integrates sensory inputs and high-level goals, transforming them into motor commands through various pathways. These pathways can be either centralized—comprising the ascending path to the brain and the descending path from the brain to the muscles—or local, involving the reflex arc path between the muscles and the spinal cord. Given the diverse operational pathways, the human nervous system can be conceptualized as a hierarchical control system, with multiple loops running in parallel \cite{binder2009} \cite{rapetti2023control}. This conceptualization has led to the proposal of a hierarchical human motor control model, which mirrors the three-layer control architecture used in humanoid robots, as shown in Figure\ref{fig:diagram_robot_parametrization}c. In this model, human behavior results from multiple loops operating concurrently. The mathematical models that describe these phenomena can be simplified into motor control models. In the proposed representation, the hierarchical nature of motor control, analogous to that of humanoid robots, includes three key components: a trajectory planning layer responsible for translating objectives into whole-body trajectories, a trajectory adjustment layer that fine-tunes these trajectories based on sensory feedback and reduced models, and a trajectory control block that generates control inputs for the actuators/muscles. The model of the human central nervous system shares these key components with the humanoid robot's physical intelligence while, the peripheral nervous system model manages local actions, such as muscle activity at the joints and sensory processing. This symmetry between human and robot physical intelligence enables seamless integration of the human model into the humanoid robot control architecture.
\subsubsection{Parametrized human-robot coupled dynamics}
\label{sec:methods:robot-load-coupled-dynamics}
The proposed framework requires a description of the dynamics of the whole system accounting for the interaction among multiple agents.
For the sake of simplicity, and to guarantee the feasibility of experimental validation, we will restrict the discussion to the case where two agents manipulate a single rigid body object, as represented in Figure \ref{fig:diagram_robot_parametrization}.  In this section, we will use the subscript $R$, $H$, $o$ to refer, respectively, to the robot, human, and object quantities.
The dynamics of each agent, considered in isolation, is described by dynamics description in Equation \eqref{eq:constrained-dynamic-hardware-param:dynamics} and the holonomic constraints in Equation \eqref{eq:constrained-dynamic-hardware-param:constraints}, with or without the parametrization w.r.t. hardware parameters. During physical collaboration, multiple contact points are present, and forces are exchanged both with the environment and between the agents and the payload. For this reason, constraints on the contact point with the environment $J^{e}\nu=0$ and those of the agent-payload contact point for each agent $J_R^{i}\nu_R=J_o^{i}\nu_o$, $J_H^{i}\nu_H=J_o^{i}\nu_o$ must be taken into account as system constraint. In addition, the  \emph{action reaction} property for internal forces must be considered, i.e.  $f^{i}_R=-f^{i}_o$ $f^{i}_H=-f^{i}_o$. This leads to the definition of the following constraint dynamics: 
\begin{equation}
\begin{split}
\label{eq:multi-system-equations}
& \begin{bmatrix} M_R(q_R, \pi) & 0 & 0\\\ 0 & M_H(q_H) & 0 \\ 0 & 0 & M_o(q_o)\end{bmatrix} \begin{bmatrix} \dot{\nu}_R \\ \dot{\nu}_H \\ \dot{\nu}_o \end{bmatrix} + \begin{bmatrix} h_R(q_R,\nu_R,\pi) \\ h_H(q_H,\nu_H) \\h_o(q_o, \nu_o) \end{bmatrix} =\\ & \begin{bmatrix} B_R & 0 \\\ 0 & B_H \\\ 0& 0 \end{bmatrix}  \begin{bmatrix} \tau_R \\\ \tau_H \end{bmatrix} + \mathbf{Q}(q_R,q_H,q_o, \pi)^T \mathbf{f}, \\
& {\mathbf{\dot{Q}}}(q_R,q_H,q_o,\pi) \begin{bmatrix} \nu_R \\ \nu_H \\ \nu_o \end{bmatrix} +\mathbf{Q}(q_R,q_H,q_o,\pi) \begin{bmatrix} \dot{\nu}_R \\ \dot{\nu}_H \\ \dot{\nu}_o \end{bmatrix} = 0,
\end{split}
\end{equation}
where $\mathbf{Q}$ is a coupling matrix combining contact point Jacobians, and $\mathbf{f}$ contains all the internal and external wrenches exchanged by the agents.
By identifying with bold the composite matrices, we can write Equation \eqref{eq:multi-system-equations}, in its compact form as: 
\begin{equation}
\label{eq:coupled-dynamics}
\begin{split}
& \mathbf{M} (\boldsymbol{q}, \pi){\boldsymbol{\dot{\nu}}} + \mathbf{h}(\boldsymbol{q}, \boldsymbol{\nu},\pi) = \mathbf{B} \boldsymbol{\tau} + \mathbf{Q}^T(\boldsymbol{q}, \pi) \mathbf{f}, \\
& {\mathbf{\dot{Q}}}(\boldsymbol{q}, \pi) \boldsymbol{\nu} +\mathbf{Q}(\boldsymbol{q}, \pi) \boldsymbol{\dot{\nu}} = 0.
\end{split}
\end{equation}
\subsubsection{Reduced models}\label{sec:results_centroidal_dynamics}
When considering locomotion on flat terrain (i.e. assuming unilateral planar contacts with no slipping nor tilting), the full floating-base parametrized dynamics is not necessary for control design, and template models suffice. Typically, those models are derived using Newton-Euler balance of forces and moments equations. One such model is Orin's centroidal dynamics; equations describing the evolution in time of the aggregate linear and angular momentum of all links referred to the robot center of mass - CoM, and oriented as the inertial frame due to contact forces and other external perturbations \cite{orin}. These equations read as:
\begin{subequations}\label{centroidal_dynamics_eqns}
\begin{align}
\dot{p}_{CoM} &= \frac{1}{m} C {}_G h\label{com_dynamics}\\
{}_G \dot{h} &= m g + \sum_{i = 1}^{n_c}\begin{bmatrix} \mathbb{I}_3 & 0_3 \\ (p_i - p_{CoM})^{\wedge}  & \mathbb{I}_3\end{bmatrix} \  f_i  + \sum_{i = 1}^{n_d} \begin{bmatrix} \mathbb{I}_3 & 0_3 \\ ( p_{d_i} - p_{CoM})^{\wedge}  & \mathbb{I}_3\end{bmatrix} \  d_i \label{momentum_dynamics}
\end{align}
\end{subequations}
in which Equation (\ref{com_dynamics}) describes the dynamics of the CoM position as a function of the robot mass $m$ and its aggregate  momentum ${}_G h$ through a selector matrix $C = \begin{bmatrix}
    \mathbb{I}_3 & 0_3
\end{bmatrix}$. Similarly, Equation (\ref{momentum_dynamics}) describes the dynamics of the aggregate momentum ${}_G h$ as a function of the 6D gravity vector $g$, and the balance of external contact forces $f_i$ and perturbations $d_i$, and their locations $p_i, \ p_{d_i}$ respectively. In addition, to simulate whether a given  contact force at the feet $f_i$ is \textit{active} or not during stepping, and following \cite{Romualdi2022}, we considered the contact locations $p_i$ as continuous variables with the following dynamics
\begin{equation}\label{contact_pos_dynamics}
    \begin{aligned}
        \dot{p}_i &= (1-\gamma_i) v_i , \qquad \gamma_i \in \{ 0, \ 1\}
    \end{aligned}
\end{equation}
    where $v_i$ is the contact velocity.
\subsection{Optimization of humanoid robot design  for ergonomic human-robot collaboration}
\label{sec:methods:morphology_optimization}
In this section, we present details about the implementation of the architecture depicted in Figure \ref{fig:istances}a, which results in the ergoCub optimal hardware, as presented in Section \ref{sec:results:ergocubRobot}.
To identify the morphology of the ergoCub robot, we formulated the optimization problem as per  Equation \eqref{eq:optimizationProblem}. 
\begin{IEEEeqnarray}{RCL}
\IEEEyesnumber
\label{eq:optimizationProblem}
\boldsymbol{y} ^{\star} & = & \underset{\boldsymbol{y}}{\text{argmin}}\left(    W_1 T_1{+}W_2 T_2{+}W_3 T_3\right)  \nonumber\\
& s.t. & \nonumber\\ 
&& \mathbf{M} (\boldsymbol{q}, \pi){\boldsymbol{\dot{\nu}}} + \mathbf{h}(\boldsymbol{q}, \boldsymbol{\nu},\pi) = \mathbf{B} \boldsymbol{\tau} + \mathbf{Q}^T(\boldsymbol{q} , \pi) \mathbf{f},  \IEEEyessubnumber  \label{coupled_dynamic} \\
&& {\mathbf{\dot{Q}}}(\boldsymbol{q}, \pi) \boldsymbol{\nu} +\mathbf{Q}(\boldsymbol{q}, \pi) \boldsymbol{\dot{\nu}} = 0,  \IEEEyessubnumber \label{holonomic_contraints}\\
&& H_i = H_i^*, \IEEEyessubnumber \label{task_constraints}\\
&& \boldsymbol{\dot{\nu}} = 0,  \IEEEyessubnumber  \label{static_conditions_1}\\
&& \boldsymbol{\nu} = 0.  \IEEEyessubnumber   \label{static_conditions_2}\\
\nonumber
\end{IEEEeqnarray}
The optimization problem search variables is defined as  $\boldsymbol{y} =     \begin{bmatrix}
\tilde{\boldsymbol{q}} &
\pi
\end{bmatrix}^T$ where $\pi$ are the robot link lengths, and $\tilde{\boldsymbol{q}}$ are the human and robot configurations associated with different load heights. $\tilde{\cdot}$ represents a set of variables.
In this optimization problem, we leveraged the parametrized coupled dynamics of human-robot-load of Equation \eqref{eq:coupled-dynamics}, but considering only the robot link lengths as hardware parameters, thus disregarding the link densities. Following the same convention of Section \ref{sec:methods:robot-load-coupled-dynamics}, we identify the agents configuration with $\boldsymbol{q}$ and the agents' configuration velocity with $\boldsymbol{\nu}$. The interaction wrenches are identified with  $\boldsymbol{f}$ and with $\boldsymbol{Q}$ we defined the coupling matrix that considers both the constraints of the contact with the environment and those exerted in between the agents. 
The human and robot internal joint torques $\boldsymbol{\tau}$ are incorporated into the system dynamics via a constant selector matrix $\boldsymbol{B}$
meanwhile with $\mathbf{M}$ we refer to the coupled mass matrix and $\mathbf{h}$ accounts for the Coriolis and gravity effects.
In the formulation of the optimization problem, in addition to the coupled dynamics of Equation \eqref{coupled_dynamic}, we integrated several constraints to ensure task feasibility. These encompass kinematic constraints (Equation \eqref{task_constraints}) to maintain the correct load orientation and height, as well as to position the hands and feet appropriately for both agents during the task. Additionally, we impose  contact constraints (Equation \eqref{holonomic_contraints}) and static conditions (Equation \eqref{static_conditions_1}.
Our goal was to find the set of optimum robot link lengths $\pi^*$ to improve collaborative and walking task performances. For this reason, we formulated three tasks weighted by $W_1, W_2, W_3 \in \mathbb{R}^+$. For the ergoCub optimization, the chosen values were $W_1 = 10^5$, $W_2 = 10^{10}$, and $W_3 = 1$. To improve the walking performances we lifted the robot center of mass by adding in the cost function the task $T_3 = \left\rVert\frac{1}{\prescript{\mathcal{I}}{}{p}_{z0,com}}\right\rVert_2^2$ where $\prescript{\mathcal{I}}{}{p}_{z0,com}$ is the robot center of mass height computed at null configuration.
To improve the human-robot collaboration, we improve the ergonomy of the interaction in the form of agents energy expenditure. A measure of the agents energy expenditure is the norm of both agent joint torque. For this reason, we add two tasks minimizing the agents joint torque, i.e. $T_1=\left\rVert S_R{\boldsymbol{\tau}} (\boldsymbol{q}, \pi)\right\rVert_2^2 $ and $T_2=\left\rVert S_H\boldsymbol{\tau} (\boldsymbol{q}, \pi)\right\rVert_2^2 $, with $S_R, S_H$ are selector matrix for robot and human torque respectively. $\boldsymbol{\tau} (\boldsymbol{q}, \pi)$ is computed by projecting the dynamics of Equation \eqref{eq:coupled-dynamics} into the holonomic constraints, thus obtaining Equation \eqref{eq:constrained-dynamic-projected}  with $\boldsymbol{N}_{\lambda}(\boldsymbol{q},\pi)$ defined as Equation \eqref{eq:N_A} (where the dependency on $\boldsymbol{q}$ and $\pi$ has been omitted for the sake of conciseness). 
\begin{align}
\label{eq:constrained-dynamic-projected}
\begin{split}
& \boldsymbol{N}_{\Lambda}(\boldsymbol{q}, \pi) [\boldsymbol{M}(\boldsymbol{q}, \pi)\dot{\boldsymbol{\nu}} + \boldsymbol{h}(\boldsymbol{q}, \pi,\boldsymbol{\nu})-B\boldsymbol{\tau}]=0,
\end{split}
\end{align}
\begin{equation}
\label{eq:N_A}
\boldsymbol{N}_{\Lambda}=1-\boldsymbol{Q}^T\left(\boldsymbol{Q} \boldsymbol{M}^{-1}\boldsymbol{Q}^T\right)^{-1}\boldsymbol{Q} \boldsymbol{M}^{-1}.
\end{equation}
Equation \eqref{eq:constrained-dynamic-projected}, is then used to express the relationship between the agents joint torques and the search variable $y=\begin{bmatrix}
\tilde{\boldsymbol{q}} &
\pi
\end{bmatrix}^T$ at static configurations and considering a minimum norm solution, leading to the following relationship: $\boldsymbol{\tau}(y) =\left(\boldsymbol{N}_{\Lambda}(\boldsymbol{q}, \pi)\boldsymbol{B}\right)^{\dagger}\boldsymbol{N}_{\Lambda}(\boldsymbol{q}, \pi)\boldsymbol{g}(\boldsymbol{q}, \pi)$, with $\boldsymbol{g}(\boldsymbol{q},\pi)$ accounting for the gravity effects only. This computation enables to constraint the optimization problem to the parametrized couple dynamics without explicitly incorporating it into the optimization problem constraints, avoiding potential computational issues.
This torque relationship is also used in the analysis of the optimization outcomes presented in Section \ref{sec:results:ergocubRobot}. Specifically, the human joint torque values reported in Figure \ref{fig:hardware_optimization_output} (e) are computed using this formulation.
\subsection{Physical intelligence}
\label{sec:methods:shared_control_framework}
In this section, we present details about the implementation of the control framework used in the architecture proposed in Section \ref{sec:results:ergoCub_physical_intelligence}. In such an architecture instance, we developed trajectory control, adjustment, and planning blocks that consider the influence of a human companion while keeping the hardware parameters fixed to the optimal values obtained in the previous instance. The ultimate goal is to generate instances of the trajectory blocks that incorporate human representation and achieve partner awareness.
Starting from the dynamics of Equation \eqref{eq:coupled-dynamics}, it is possible to derive a set of control inputs $u^*$, i.e. joint torques $\boldsymbol{\tau}$ and contact forces $\mathbf{f}$, defining trajectory blocks, able to achieve the desired task while minimizing both human and robot metrics, and considering the optimum set of hardware parameters $\pi^*$.
\begin{IEEEeqnarray}{RCL}
\label{eq:optimization_trajectory_control}
\IEEEyesnumber
u^{*} & = & \underset{u}{\text{argmin}}\left( \boldsymbol{\Psi}_{u}(u) \right)   \nonumber\\
& s.t. & \nonumber\\ 
&&  \mathbf{M} (\boldsymbol{q} | \pi^{*}){\boldsymbol{\dot{\nu}}} + \mathbf{h}(\boldsymbol{q}, \boldsymbol{\nu} | \pi^{*}) = \mathbf{B} \boldsymbol{\tau} + \mathbf{Q}^T(\boldsymbol{q} | \pi^{*}) \mathbf{f} \nonumber\\
&& {\mathbf{\dot{Q}}}(\boldsymbol{q} | \pi^{*}) \boldsymbol{\nu} +\mathbf{Q}(\boldsymbol{q} | \pi^{*}) \boldsymbol{\dot{\nu}} = 0
\end{IEEEeqnarray}
In addition, the same architecture can be used, disregarding the human, to achieve other behaviors, like walking robustly, as exemplified in the ergoCub robot physical intelligence, presented in Section \ref{sec:results:ergoCub_physical_intelligence}.
In the following section, we will delve into the definition of each block that composes the architecture physical intelligence, which is organized as a cascade control architecture, hence composed of the \emph{trajectory control}, \emph{adjustment}, and \emph{planning} blocks. We will delve into both the collaborative and locomotion control architectures. 
The proposed architectures builds upon prior work conducted on the iCub\cite{natale2017} and iCub3\cite{dafarra2024icub3} humanoid robots. 
Collaborative controllers developed for the iCub facilitated human-robot collaboration with the concept of assistance \cite{Tirupachuri2020} and allowed for robot-robot collaborative lifting \cite{rapetti2021}. With the iCub3, we incorporated the modeling of the human balancing system into the collaborative controller \cite{rapetti2023control}. Regarding locomotion controllers, the initial architecture designed for the iCub employed predictive control on the robot Zero Moment Point (ZMP) \cite{dafarra2018}, which subsequently evolved with the application of Centroidal Model Predictive Control (MPC) on the iCub3 robot \cite{Romualdi2022, Elobaid_parametrizedCentroidal}.
With the introduction of the ergoCub robot, we have further refined and adapted these controllers to suit the ergoCub-specific features that arise from the performed hardware optimization.    In the following section, we will see in detail the control architectures employed for performing human-robot collaboration (Section \ref{sec:method:human-robot-coll}) and locomotion (Section \ref{sec:method:locomotion}) on the ergoCub humanoid robot. 
\subsubsection{   Human-robot collaboration-payload lifting task}
\label{sec:method:human-robot-coll}
   In this Section, we will hover over the main blocks that compose the human-robot collaboration architecture visually depicted in Figure \ref{fig:diagram_robot_parametrization}(c). The description of the employed control framework can be decomposed into the description of \textit{trajectory control}, \textit{trajectory adjustment}, and \textit{trajectory planning} blocks.
\noindent{\textbf{Trajectory control}}
\noindent The trajectory control represents the inner loop of the control framework preceding the low-level control of the actuators. It demands both a high frequency and a comprehensive system model, consequently, predictive horizons need to be sacrificed while instantaneous control is employed.
Given reference robot joints trajectory $(s^r_R, \dot{s}^r_R, \ddot{s}^r_R)$ or task-space robot reference trajectories $(x^r_{R,n}, \dot{x}^r_{R,n}, \ddot{x}^r_{R,n})$, it is required to design a control law that defines the actuation command based on the dynamics model of the system. 
In the human-robot collaboration task,    the robot is actuated through joint torques $\tau_R$, in this way, it can be \emph{compliant}, as usually is the case for Human-Robot interaction  \cite{hyon2007full}. We employed a task-space inverse dynamics approach \cite{delprete2015}, together with the coupled human-robot dynamics of Equation \eqref{eq:coupled-dynamics}, thus considering the human-robot mutual influence, and we formulate the trajectory control as a linear constrained optimization problem:
\begin{equation}
\label{eq:tsid}
    \begin{aligned}
    \min\limits_{\boldsymbol{\tau}} \quad & \left \| \ddot{s}_{R} - \ddot{s}_{R}^{*} \right \| ^2 + \sum_{n = 0}^{N-1} \left \| \ddot{x}_{R,n} - \ddot{x}_{R,n}^{*} \right \| ^2  \\
         s.t. \quad & \mathbf{M} (\boldsymbol{q}| \pi^*){\boldsymbol{\dot{\nu}}} + \mathbf{h}(\boldsymbol{q}, \boldsymbol{\nu}|\pi^*) = \mathbf{B} \boldsymbol{\tau} + \mathbf{Q}^T(\boldsymbol{q}|\pi^*) \mathbf{f}, \\
& {\mathbf{\dot{Q}}}(\boldsymbol{q}| \pi^*) \boldsymbol{\nu} +\mathbf{Q}(\boldsymbol{q}| \pi^*) \boldsymbol{\dot{\nu}} = 0, \\
& {\mathbf{\dot{J}}_n}(\boldsymbol{q}| \pi^*) \boldsymbol{\nu} +\mathbf{J}_n(\boldsymbol{q}| \pi^*) \boldsymbol{\dot{\nu}} = \ddot{x}_{R,n}^{*}, \\
& n = 0,1, ...,N-1.
    \end{aligned}
\end{equation}
where $\pi^*$ is the optimal set of hardware parameters, identified as explained in Section \ref{sec:methods:morphology_optimization}    and the subscript $R$ highlights that the quantities are robot-related. 
To achieve task-space tracking, we use the following:
\begin{equation}
\label{eq:task-space-tracking}
    \begin{aligned}
    \ddot{x}_{R,n}^{*} = \ddot{x}^r_{R,n} + K_d(\dot{x}^r_{R,n} - \dot{x}_{R,n}) + K_p ({x}^r_{R,n} -    {x}_{R,n}) + K_I \int ({x}^r_{R,n} - {x}_{R,n}).
    \end{aligned}
\end{equation}
Similarly, for joint tracking, we use:
\begin{equation}
\label{eq:joint-tracking}
    \begin{aligned}
    \ddot{s}_{R}^{*} = \ddot{s}^r_R + k_d(\dot{s}^r_R - \dot{s}_R) + k_p ({s}^r_R - s_R).
    \end{aligned}
\end{equation}
These equations ensure the tracking of desired tasks when $K_d$, $K_p$,$K_I$, $k_d$, and $k_p$ are diagonal positive definite matrices. The constrained optimization problem in Equation \eqref{eq:tsid} involves both quadratic and linear functions, enabling efficient online solutions at high frequencies by exploting QP.
\noindent{   \textbf{Trajectory adjustment}}
\noindent
While tracking a pre-computed trajectory can enable the achievement of simple tasks in controlled environments, the presence of collaborative agents necessitates the adjustment of the trajectory for dynamical adaptation to the surroundings. In this case, we are interested in adjusting the time evolution of the planned trajectory, while we are not modifying the trajectory \textit{per se}. We have previously explored the idea of parametrizing the trajectory for human-robot collaboration in \cite{Tirupachuri2019} and \cite{rapetti2023control}. Starting from the latter, we proposed an updated version of trajectory parametrization for collaborative lifting, which further promotes ergonomics collaboration.    The reference robot configuration $s^r_R(t)$ for determining the motor control input of the trajectory control block, as per Equation \eqref{eq:tsid},  is computed using Equation \eqref{eq:trajectory-parametrization}.
\begin{equation}
    \label{eq:trajectory-parametrization}
    \begin{aligned}
        s^r_R(t) = s_{R}(\psi(t)) = s_1 + \int_0^t \dot{s}_{   R}(\psi(t)) dt, \\
        \dot{s}^r_{   R}(\psi(t)) = (s_2 - s_1) \dot{\psi}(t).
    \end{aligned}
\end{equation}
   Therefore, $s^r_R$ is define to move from a joint configuration $s_1$ to $s_2$ and it is regulated by a free parameter, $\psi = \psi(t) \in [0, 1]$.
To regulate the evolution of the free-parameter, we need to define a task-specific quantity that acts as a master. In the case of collaborative lifting, this corresponds to the motion of the human hands which should be followed by the robot motion. In synthesis, we obtain a desired robot hands velocity and we use it to regulate the free parameter:
\begin{equation}
  \label{eq:free-parameter-evolution}
  \begin{aligned}
      \dot{z}_R = k_{\psi} \cdot  (z_H -  z_R), \\
      \dot{\psi} = \left [ J^z_s \cdot (s_2 - s_1) \right ]^{\dagger} \dot{z}_R,
  \end{aligned}
\end{equation}
where $z_H$ and $z_R$ are respectively the height of the human and the robot hand, $J^z_s$ is the component of the robot hand jacobian associated with the velocity along the z-axis, and $k_{\psi} \in \mathbb{R}^{+}$ is a gain that determines the robot reactivity in following human hand motion.    The robot, therefore, will be governed by the human action as per Equation \eqref{eq:free-parameter-evolution} only inside the range $s_1,s_2$ meanwhile it will not move outside such a range. 
\noindent\textbf{   Trajectory planning}
\noindent
   In trajectory planning, starting from the desired height range the load should be manipulated, we computed two references initial $s_1$ and final $s_2$ joint configuration, solving an Inverse Kinematic problem. These reference joint positions, $s_1$ and $s_2$, are then employed in the trajectory adjustment block, as described in Equation  \eqref{eq:free-parameter-evolution} . 
\subsubsection{   Locomotion}
\label{sec:method:locomotion}
   Also for the locomotion control, we employed a hierarchical control architecture defined by the trajectory control, adjustment and planning blocks. 
\noindent\textbf{   Trajectory control}
\noindent
   As is usual by now, this block is aimed at providing direct inputs to the low level actuators control. 
For the locomotion behavior, we have two different trajectory blocks, depending if the low level input are joint torques or positions. 
   When the input is joint positions, as for the ergoCub robot locomotion instance, the reference joint positions $s^r_R$ are evaluated through a QP-based inverse kinematics module and they are passed to the actuator low-level motor control boards \cite{Giulio_benchmarking}. In this instance, a \textit{stack of tasks} is employed in which tracking of the feet poses and the CoM
position is treated as a \textit{hard} constraint,
while the torso orientation is treated as a cost term. In this sense, at each time step one solves:
\begin{equation}\label{ik_qp}
    \begin{aligned}
        \min\limits_{\boldsymbol{\nu}_{   R}} \quad &\frac{1}{2} \nu_{   R}^{\top} H \nu_R + g^{\top} \nu_{   R}\\
        s.t. \quad & A\nu_R = b \\
        \quad & \dot s_{{   R}},\ell \leq \dot s_{   R} \leq \dot s_{R,u}
    \end{aligned}
\end{equation}
where in this formulation, the optimization variable is the robot velocity $\nu_R$, and the feedback is obtained by integrating the desired quantities. The hessian matrix and gradient vector  $H, \ g$ are obtained from the torso and joints regularization tasks respectively, while the constraint matrices $A, \ b$ and robot joint velocity lower and upper limits $\dot s_{R,\ell}, \ \dot s_{R.u}$ are obtained from the jacobians associated with the velocity of the CoM and feet, together with actuator limits respectively. 
\noindent\textbf{   Trajectory adjustment}
\noindent
   Also in the locomotion case, a trajectory adjustment block is needed, since complex tasks — such as creating and removing contacts — necessitate the adjustment of the trajectory for dynamical adaptation to the surroundings. To achieve this, simplified models with lower computational costs are employed, eventually allowing for the inclusion of a prediction horizon.
In our case we utilized Model Predictive Control (MPC) which allows finding the system state $\boldsymbol{x}$ and control input $\boldsymbol{u}$ over a defined time horizon. In \cite{Romualdi2022}, it was proposed the centroidal MPC, which can adjust online the trajectory of a humanoid robot during locomotion, and further improved in \cite{Elobaid_parametrizedCentroidal} for locomotion with payloads. 
The tasks considered in the loss function are the position of the contact points, rate of change of the feet impact wrenches, angular momentum tracking, and symmetry in the feet impact wrenches. More in detail, and starting from the dynamical system (Equation \eqref{centroidal_dynamics_eqns}), we aim to adjust the feet contact locations and timings solving the following optimization problem via receding horizon \cite{grune}
over $n_p \geq 1$ future time steps;
\begin{subequations}\label{centroidal_mpc_with_payload}
\begin{align}
\underset{f_i, \ v_i}{\text{minimize}} \ \ \sum_{i = 1}^{n_p} &W_{{}_G h} hT_{{}_G h}[k+i] + \ W_fT_f[k+i] + \ W_pT_p[k+i] \label{cost}\\
& s.t \nonumber \\
{}_G h[k+1] &= {}_G h[k] + \Delta T\left( m\Vec{g} + \sum_{i = 1}^{n_c}\begin{bmatrix} \mathbb{I}_3 & \mathbf{0}_3 \\ (p_i - p_{CoM})^{\wedge} & \mathbb{I}_3 \end{bmatrix} \ f_i\right) \nonumber \\ & +  \sum_{i = 1}^{n_d}\begin{bmatrix} \mathbb{I}_3 & \mathbf{0}_3\\ (p_{d_i} - p_{CoM})^{\wedge} & \mathbb{I}_3 \end{bmatrix} \ d_i   \label{dyn_1} \\
p_{CoM}[k+1] &= p_{CoM}[k] + \Delta T\left(\frac{1}{m}C{}_G h\right) \label{dyn_2}\\
p_i[k+1] &=  p_i[k]  + \Delta T ((1 - \gamma_i) \ v_i) \label{dyn_3} \\
 f_i &\in  \mathcal{K}_i \label{contact_force_constraints} \\
\ell_{b} & \leq R_{_{i}p}^\top (p_i - p_i^n) \leq u_b \label{contact_pos_constraint}
\end{align}
\end{subequations}
where at the current time step $k$ and with a sampling period $\Delta T$ we have;
\begin{itemize}
    \item Equation (\ref{cost}) describes the function to be minimized, comprising: $T_{{}_G h}$ a term penalizing the ${L}_2$-norm of the error between the desired and current CoM position $p_{CoM}$ and aggregate momentum. Similarly $T_{f}, \ T_p$ regularizes the contact forces at the feet vertices ($f_i$) to be as similar as possible and minimizes the tracking error of the footsteps contact locations $p_i$ respectively. $W_{{}_G h}, W_f,W_p \in \mathcal{R}^{+}$ are weights.
    \item Equation (\ref{dyn_1}-\ref{dyn_3}) describes the Euler discretization of the centroidal dynamics and contact position dynamics described in (\ref{centroidal_dynamics_eqns})-(\ref{contact_pos_dynamics}) respectively.
    \item Equation (\ref{contact_force_constraints} - \ref{contact_pos_constraint}) describe the constraints restricting the obtained optimal co-planar contact forces $f_i$ to be \textit{feasible} \cite{gabri_parametrization}, i.e. in the sense of lying in the interior of the friction cone, and the maximum allowable contact location tracking error within given bounds $\ell_b, \ u_b$ respectively.
\end{itemize} 
The outputs of the above are: adjusted (given feedback and estimates of the presence of disturbances) trajectories for the CoM, the footsteps locations, and their timings. This is then passed to the whole-body trajectory control layer.
\noindent   \textbf{Trajectory planning}
\noindent
   For the walking task, the trajectory planning block generates nominal footstep and CoM trajectories for the centroidal MPC block. Such references are generated either starting from user inputs, given via a remote controller and employing an optimal control based trajectory planner (https://github.com/robotology/unicycle-footstep-planner).
\graphicspath{{Figures/}} 
\begin{appendices}
\renewcommand{\theequation}{A\arabic{equation}}
\section{Collaborative walking with payload}
\label{sec:collaborative-walking}
To further validate the ergoCub hardware design in human–robot collaboration, we performed an additional experiment in which a human and the robot walk together while carrying a payload, as illustrated in Figure \ref{fig:CollWalking} (a). The control architecture was the same as that used for locomotion in Section 2.2, with the addition of two constraints: i) the robot was commanded to maintain a constant distance from the human, and ii) an inverse kinematics task was added to ensure that the robot’s hand maintained contact with the payload. It is important to note that in this setup, the robot’s physical intelligence does not incorporate human ergonomic metrics or an explicit representation of the human partner. The purpose of these experiments is instead to assess the suitability of the robot’s hardware design for collaborative interaction, given that the design was optimized with respect to both walking and human–robot collaboration tasks.  In the experiments, the human and robot jointly carried a \SI{2}{kg} payload in two trials lasting 130 and 160 seconds, respectively Which is 4 times longer than the collaborative payload lifting experiments. Additionally, these experiments were conducted with a different operator than in the collaborative lifting study. The human participant wore the iFeel suit to capture ergonomic data. Figure \ref{fig:CollWalking} (c) reports the measured torques at the human lower back (L5S1), computed using the same methodology described in Section 2.2. The signals were smoothed using a moving average filter with a window size of 20. Figure \ref{fig:CollWalking} (d)  summarizes the mean and maximum torque values observed during the two tasks. The mean torque was consistent across both trials (approximately 14 Nm), while the maximum values differed slightly. These peaks are likely due to transient stress events (e.g., momentary misalignment of the payload), which is expected since the robot did not actively regulate ergonomics in this scenario, and the payload was grasped only through an additional task in the inverse kinematics. As expected, the maximum torque recorded in these experiments was higher than in the collaborative payload lifting of Section 2.2. However, it is important to account for inter-subject variability: in this experiment, the human participant was taller (180~cm vs. 173~cm) and heavier (80~kg vs. 65~kg) than the subject in the lifting experiments, which naturally shifts the threshold for critical back stress. Overall, these results indicate that collaborative walking with payload is feasible, and the recorded ergonomic metrics show good repeatability (comparable mean L5S1 torque across two trials) even in longer-duration interactions. This experiment complements the collaborative lifting study by validating that the ergoCub hardware design supports robust and repeatable human–robot collaboration across different tasks, durations, and human partners.
\renewcommand{\theequation}{B\arabic{equation}}
\setcounter{equation}{0}
\section{Comparison between iCub3 and ergoCub for human–robot collaboration}
\label{sup:comarison-human-robot-coll}
To further validate the proposed framework, we extend our analysis of the human–robot collaboration task by comparing the performance of the ergoCub robot with that of iCub3. Indeed, while both are humanoid robots, iCub3 was not optimized for collaborative payload lifting, although it has been used for such tasks, as reported in \cite{rapetti2023control}. Unfortunately, detailed information on human ergonomic metrics, such as internal joint stress or torque, was not recorded during the iCub3 experiments; therefore, direct ergonomic comparisons are not possible. However, a clear distinction arises in terms of operational load height range: iCub3 was tested between \SIrange{0.40}{0.75}{\meter}, whereas ergoCub has been validated across the range of \SIrange{0.8}{1.2}{\meter}. 
To assess the ergonomic value of the load range, we computed the lower back stress across multiple human models holding loads at different heights. Specifically, we considered eight different human models, defined according to the methodology in \cite{latella2019simultaneous}, representing individuals of varying sex, height, and weight. The physical characteristics of these models are summarized in Figure \ref{fig:human_models_and_stress} and are publicly available at: \url{https://github.com/robotology/human-gazebo.git}.
To assess how load height influences human effort during static load holding, we sampled a range of heights from \SIrange{0.4}{1.8}{\meter} with a step of \SI{0.05}{\meter}. For each combination load height $i$ and human model  $j \in \left \{ \text{Human}_1, \cdots \text{Human}_{8} \right \}$ we solved the optimisation problem of Equation \eqref{eq:comparison_opt}.
This problem computes the optimal configuration $q_j$ that minimizes joint torque $\tau(q_j)$ 
under static conditions (\eqref{sup_stat_1}, \eqref{sup_stat_2}), while satisfying system dynamics
(\eqref{sup_dynamic}), holonomic constraints \eqref{sup_hol} and load placement constraint \eqref{load_height}.
\begin{IEEEeqnarray}{RCL}
\IEEEyesnumber
\label{eq:comparison_opt}
{q}_j ^{\star} & = & \underset{{q_j}}{\text{argmin}}\left( \left\rVert {\tau_j} (q_j)\right\rVert_2^2\right)  \nonumber\\
& s.t. & \nonumber\\ 
&& M (q_j){{\dot{\nu_j}}} + {h_j}({q_j},{\nu_j}) = B \tau_j + J_j^T(q_j ) f_j,  \IEEEyessubnumber \label{sup_dynamic} \\
&& J(q_j) {\nu} +J(q_j) {\dot{\nu_j}} = 0, \IEEEyessubnumber \label{sup_hol}\\
&& H_{i,j} = H_{i,j}^*, \IEEEyessubnumber \label{load_height}\\
&& {\dot{\nu_j}} = 0,  \IEEEyessubnumber \label{sup_stat_1} \\
&& {\nu_j} = 0.   \IEEEyessubnumber \label{sup_stat_2} \\
\nonumber
\end{IEEEeqnarray}
As an ergonomic indicator, we extracted the average torque at the L5–S1 spinal joint, a commonly used metric for evaluating lumbar stress during lifting tasks. The results were then averaged across all eight human models to compute the mean and standard deviation at each height. Since the torque values are computed under static assumptions and serve as an indicative stress index rather than absolute measurements, we normalized all results by the maximum observed torque, producing a dimensionless stress profile across the tested height range
This allows us to identify load height intervals that are more or less ergonomically favorable, as shown in Figure \ref{fig:human_models_and_stress}.
The analysis reveals that maximum back stress occurs at the lowest tested height of \SI{0.4}{\meter}. As the height increases, the stress decreases sharply, reaching a minimum around \SI{0.8}{\meter}. Between \SI{0.8}{\meter} and \SI{1.1}{\meter}, the stress remains relatively low and stable, forming a plateau with low standard deviation. Beyond \SI{1.1}{\meter}, the stress begins to rise again, accompanied by greater inter-subject variability up to approximately \SI{1.7}{\meter}. At \SI{1.8}{\meter}, the stress slightly decreases once more together with the standard deviation, as this height corresponds to full extension for shorter individuals, while remaining within a comfortable range for taller ones. The larger standard deviation observed in the \SIrange{1.1}{1.7}{\meter} range reflects how ergonomic favorability in that region is highly dependent on individual anthropometrics. In contrast, the \SIrange{0.8}{1.2}{\meter} range exhibits both low stress and low variance, indicating a robust ergonomic spot across diverse users.
Notably, this height interval aligns precisely with the range in which ergoCub is able to collaborate effectively, as demonstrated in the main experimental results. Conversely, iCub3’s collaboration range is limited to the lower part of the spectrum (\SIrange{0.40}{0.75}{\meter}), which corresponds to the least ergonomically favorable region. It is important to note that these ranges of feasible load heights are not manually selected but are instead determined by the robot’s hardware, and in the case of ergoCub, by the outcome of the hardware optimization process. This dependence extends beyond the robot's overall height and includes the robot’s morphological proportions, such as the relative lengths of the legs, torso, and arms. For example, a robot that is tall overall but has relatively short arms or torso may struggle in reaching loads placed at medium heights. This analysis shows that hardware optimization under shared embodied intelligence principles allows ergoCub to extend collaboration capabilities while better aligning with human ergonomics across diverse users.
\renewcommand{\theequation}{C\arabic{equation}}
\setcounter{equation}{0}
\section{Comparison between iCub3 and ergoCub for locomotion}
\label{sec:comparison:ergoCub_icub3}
    We compare the capabilities of ergoCub with the iCub3 humanoid robot \cite{dafarra2024icub3}, a widely utilized platform in various applications including telexistence \cite{dafarra2024icub3}, walking \cite{Romualdi2022}, and collaborative tasks with humans \cite{rapetti2023control}, which was the starting point for the development of the ergoCub robot. The iCub3 and ergoCub robot are depicted together in Figure \ref{fig:comparison}.
   We conducted experiments campain using real hardware where both iCub3 and ergoCub robots were tested in a locomotion task simultaneously, receiving identical user inputs and implementing the same control architecture as outlined in \cite{dafarra2024icub3}. For each platform, the control feedback gains were hand-tuned separately to achieve the best locomotion performance, ensuring that the comparison focuses on morphological differences rather than being influenced by a single set of parameters shared across both robots. The goal was to assess and compare their performance in dynamic walking tasks. In Figure \ref{fig:comparison} (b), we present the trajectories of the Center of Mass (CoM) in the x and y directions during both forward and lateral walking tasks.
   The data reveals that, despite starting from the same initial position, ergoCub consistently outpaces iCub3 in both directions.  In addition, by looking at Figure \ref{fig:comparison} (c), it can be noticed that, despite the higher velocity, the ergoCub robot requires lower currents. Indeed the mean of the norm of the current of ergoCub and the standard deviation is lower with respect to the one of iCub3.  This difference in velocity can be attributed to ergoCub's optimized design and shows evidence of the ergoCub's improved capability.
   \renewcommand{\theequation}{D\arabic{equation}}
\setcounter{equation}{0}
\section{Acceptability of the ergoCub Robot}
\label{sec:Acceptability}
As introduced in Section 1, the ergoCub robot was conceived for close-proximity collaboration in industrial and healthcare domains, where ergonomics and safe physical interaction are critical. In this Appendix, we report a user study that provides complementary insight into the perceived acceptability of the resulting humanoid platform when deployed in contexts where humans and robots interact closely. In such scenarios, perceived trustworthiness and comfort can influence real-world adoption. It is worth noting that this study is not directly related to the primary focus of this work, which is the methodological framework for optimizing robot hardware and physical intelligence with respect to human-centered ergonomic metrics. Rather, it provides additional perspective on one of the tangible outcomes of the proposed framework, namely the ergoCub robot. We conducted a user study to assess the acceptability of ergoCub’s design and its simplified facial expressions, which provide ergonomic feedback during interaction (see Section 2.2). The study was run in collaboration with Ipsos (\url{https://www.ipsos.com/en}) and involved a total of 850 participants: 500 from the manufacturing sector and 350 from healthcare. Respondents were recruited via Computer-Assisted Personal Interviews (CAPI) and Computer-Assisted Web Interviews (CAWI). Each participant was asked to:
\begin{itemize}
    \item Rate the overall appeal of four robots: NAO\cite{shamsuddin2011humanoid}, Baxter\cite{Baxter}, R1\cite{R12017}, and ergoCub, on a 0–10 scale.
\item Report the emotions evoked by each of the four robot designs (positive: interest, pleasantness, joy; negative: discomfort).
\item Indicate preferences among four styles of robot facial expression: simplified, avatar, emoji, or abstract (Figure \ref{fig:comparison_Faces}).
\end{itemize}
Results in the manufacturing domain (Figure \ref{fig:comparison_Manufacture}) showed that NAO received the highest average rating (7.0), followed by ergoCub (6.7), R1 (6.0), and Baxter (5.2). Importantly, ergoCub outperformed Baxter, a widely adopted collaborative robot, and R1, which lacks locomotion and humanoid proportions. NAO elicited slightly more positive emotions (25\%) than ergoCub (23\%), but both robots showed similarly low levels of negative affect (around 10\%), suggesting broad neutrality or acceptance rather than rejection. Results in healthcare (Figure~\ref{fig:comparison_Healtcare}) followed the same trend: NAO again ranked first (6.5), followed by ergoCub (6.0), R1 (5.3), and Baxter (4.3). Absolute scores were lower across all robots, reflecting a more critical stance among healthcare professionals, but the relative ranking remained unchanged. Emotional responses mirrored the manufacturing domain, with ergoCub and NAO eliciting comparable levels of positive affect. From these findings, we conclude that although NAO was the most liked robot overall, its lack of suitability for physical collaboration means that ergoCub is the most accepted option among robots capable of direct physical interaction with humans. Baxter, despite being a benchmark collaborative platform, consistently ranked lowest, underscoring the importance of humanoid form and mobility for user acceptability.
For what concern facial expression (Figure \ref{fig:comparison_Faces}), simplified facial expressions, as implemented in ergoCub, were consistently preferred over more complex or abstract alternatives, thereby validating our design choice for ergonomic feedback. Taken together, these results indicate that ergoCub achieves a favorable balance between functional capability and user acceptability, reinforcing its potential for deployment in both industrial and healthcare settings.
\renewcommand{\theequation}{E\arabic{equation}}
\setcounter{equation}{0}
\section{Wearable technology}
\label{wereable tecnhology}
As depicted in Figure \ref{fig:human_sensors}, wearable sensors are used to update human kinematics and estimate ergonomics during the human robot interaction of Section 2.2. The setup includes: a set of IMU-based iFeel Nodes \cite{sortino2023}, which estimate the relative orientation of the limbs;  the iFeel Shoes \cite{sortino2023}, which embed Force/Torque (FT) sensors in the soles to measure ground reaction forces; and VIVE trackers (\url{https://www.vive.com/eu/accessory/tracker3/}), mounted on the feet, hands, and waist, providing absolute pose data. These sensor measurements are integrated online to estimate human kinematics and joint torques, updating the internal human representation within the robot’s physical intelligence for collaboration.
\end{appendices}

\renewcommand{\thefigure}{A\arabic{figure}}
\setcounter{figure}{0}

\begin{figure}[H]
  \centering

        \includegraphics[trim={0.0cm 8cm 0cm 0cm},clip,width=\linewidth]{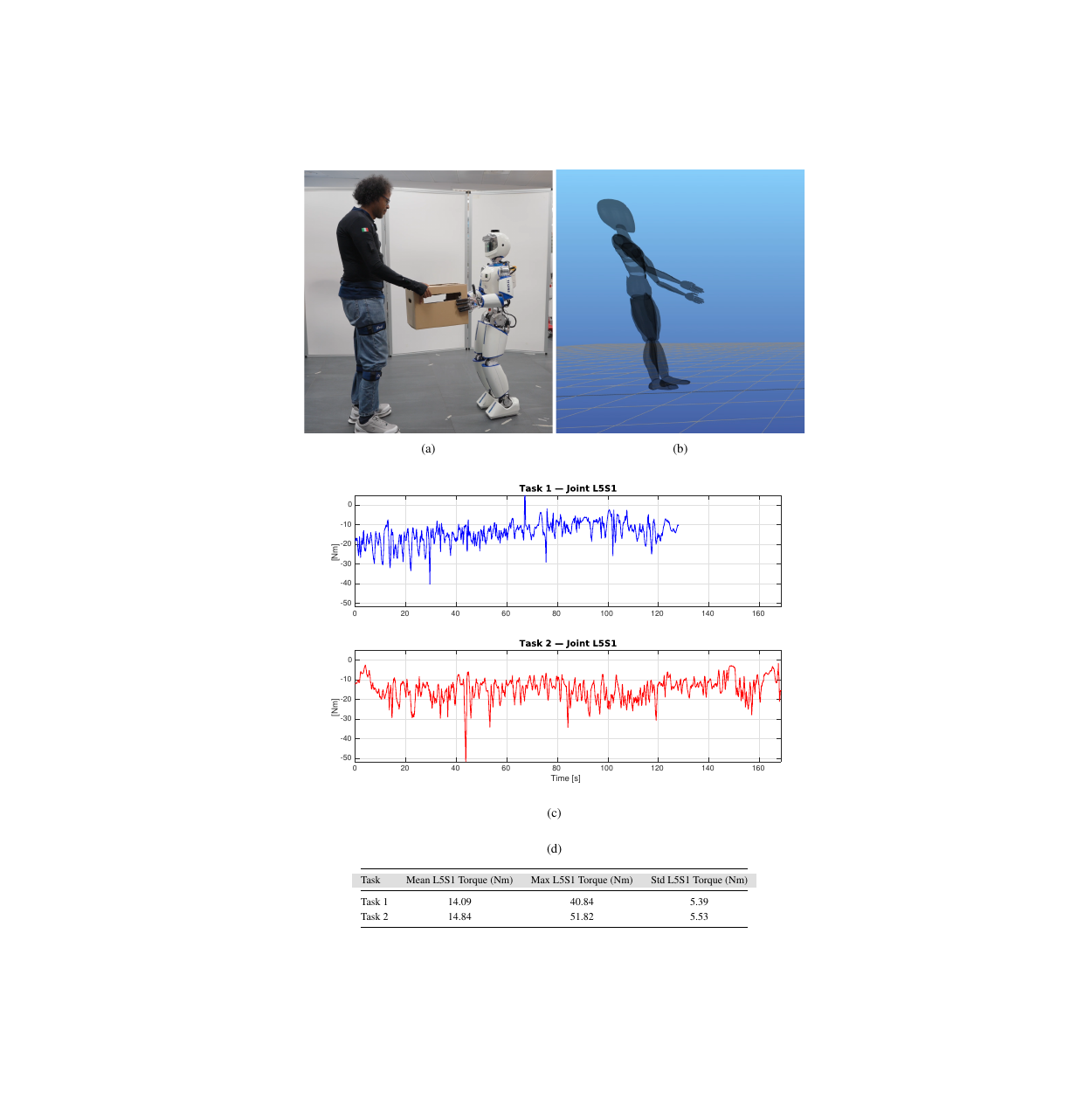}
       
  \caption{Human–robot collaborative walking with payload. (a) The ergoCub robot walking together with a human while carrying a payload. (b) Avatar representation of the human companion. (c) Estimated L5S1 joint torques during collaborative walking for the two trials. (d) Mean and maximum torque values measured on the human participant during collaboration.}
 \label{fig:CollWalking}
\end{figure}

\clearpage
\renewcommand{\thefigure}{B\arabic{figure}}
\setcounter{equation}{0}
\setcounter{figure}{0}
\clearpage
\begin{figure}[!htbp]
  
        \includegraphics[trim={0.0cm 10cm 0cm 0cm},clip,width=\linewidth]{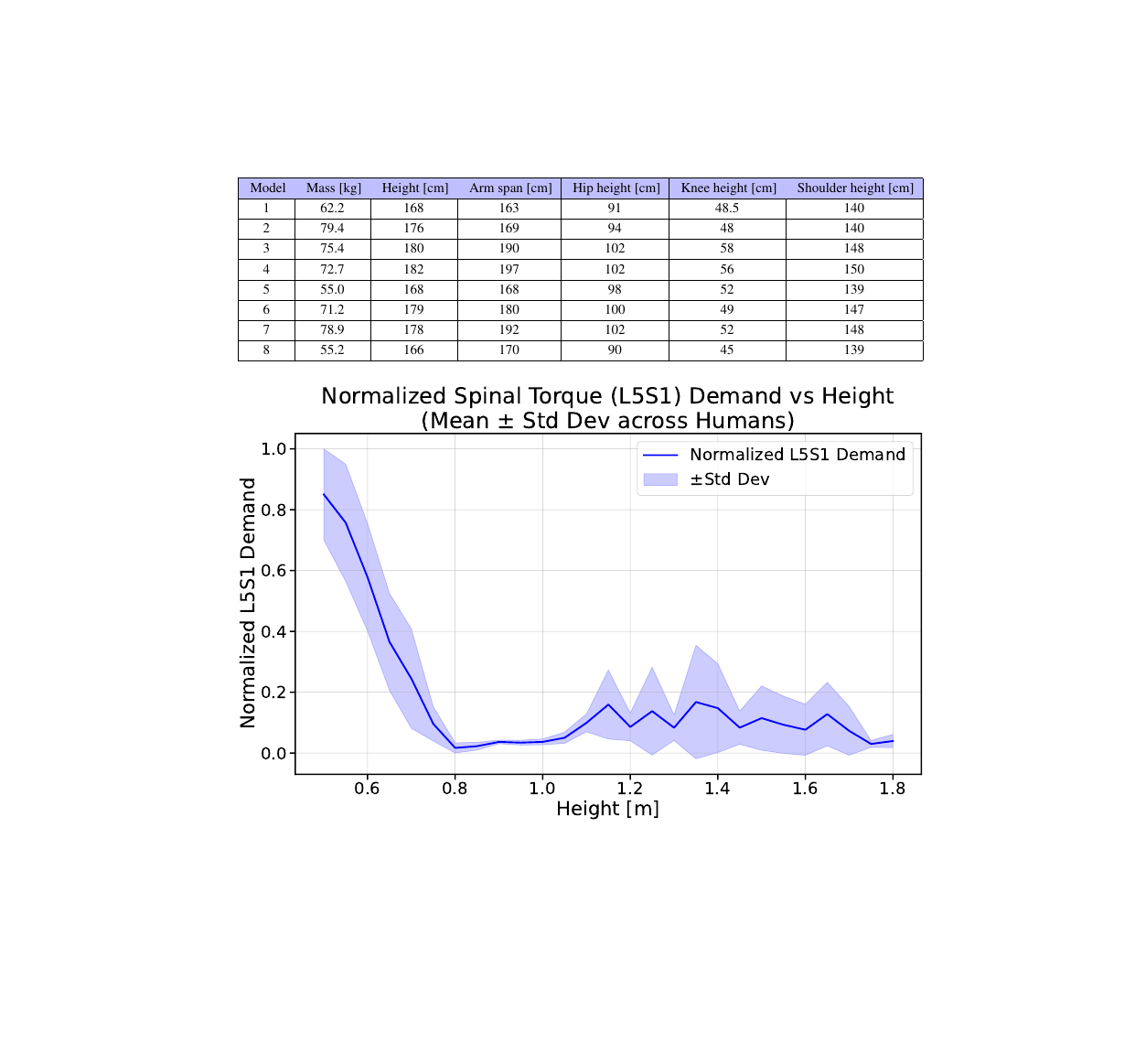}  
  \caption{Human model characteristics and normalized L5–S1 spinal torque demand as a function of load height. (a) Main characteristics of the human models used for the ergonomics analysis. 
  (b) Relative L5–S1 torque demand (mean ± standard deviation across simulated human models) as a function of target load height. 
  All values are normalized such that the maximum observed spinal torque corresponds to 1. The shaded area represents the standard deviation across the different human models.}
  \label{fig:human_models_and_stress}
\end{figure}

\renewcommand{\thefigure}{C\arabic{figure}}
\setcounter{equation}{0}
\setcounter{figure}{0}
\clearpage
\begin{figure}[H]
  \centering
  
        \includegraphics[trim={0.0cm 8cm 0cm 0cm},clip,width=\linewidth]{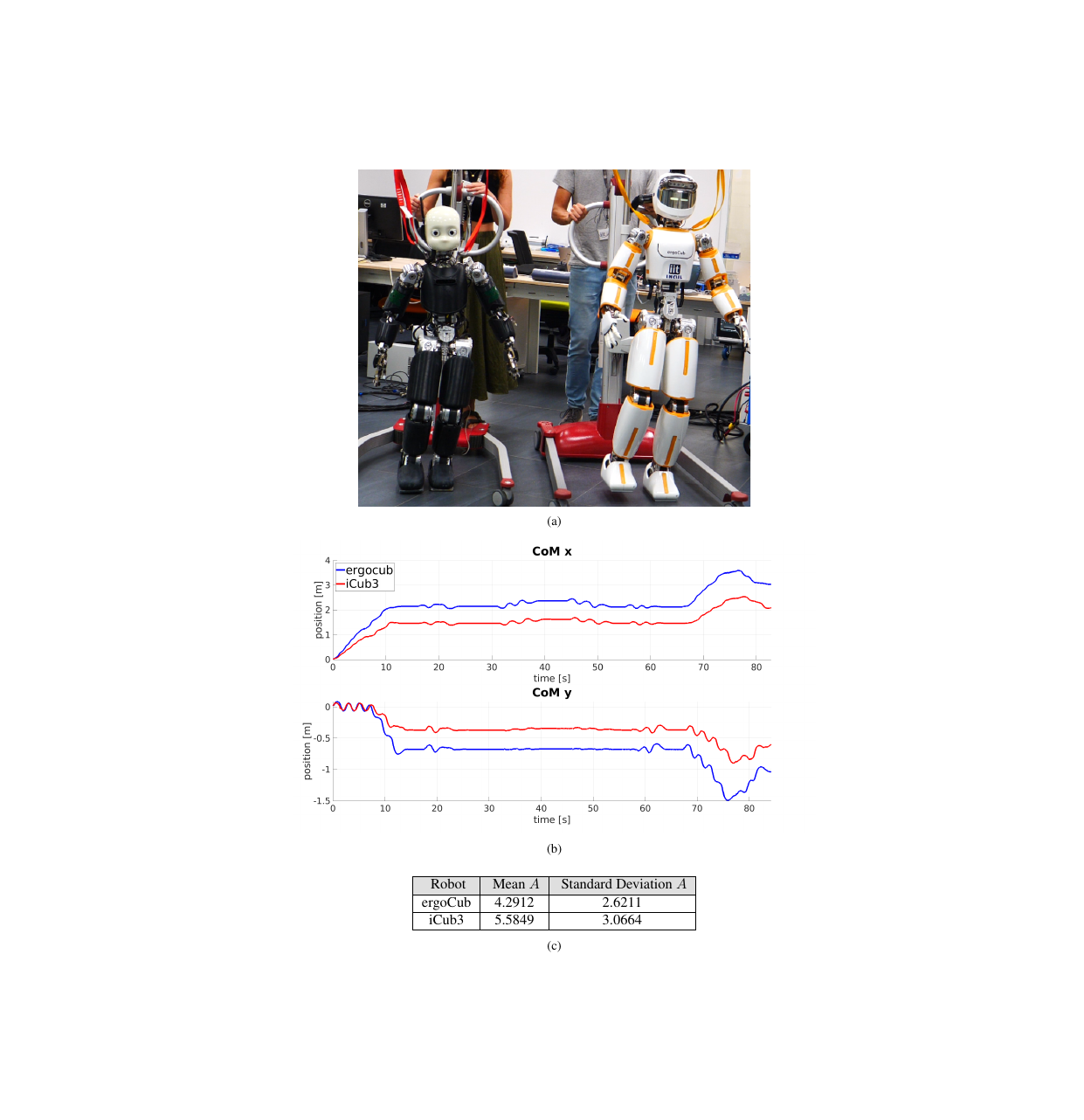}  
  \caption{Comparison between the ergoCub robot and iCub3. The ergoCub and iCub3 robots walking together (a). Comparison of Center of Mass (CoM) trajectories for both robots (b). Mean and standard deviation of the current norm during the walking task for ergoCub and iCub3 (c).}
  \label{fig:comparison}
\end{figure}

\renewcommand{\thefigure}{D\arabic{figure}}

\setcounter{equation}{0}
\setcounter{figure}{0}
\clearpage
\begin{figure}[H]

        \includegraphics[trim={0.0cm 10cm 0cm 0cm},clip,width=\linewidth]{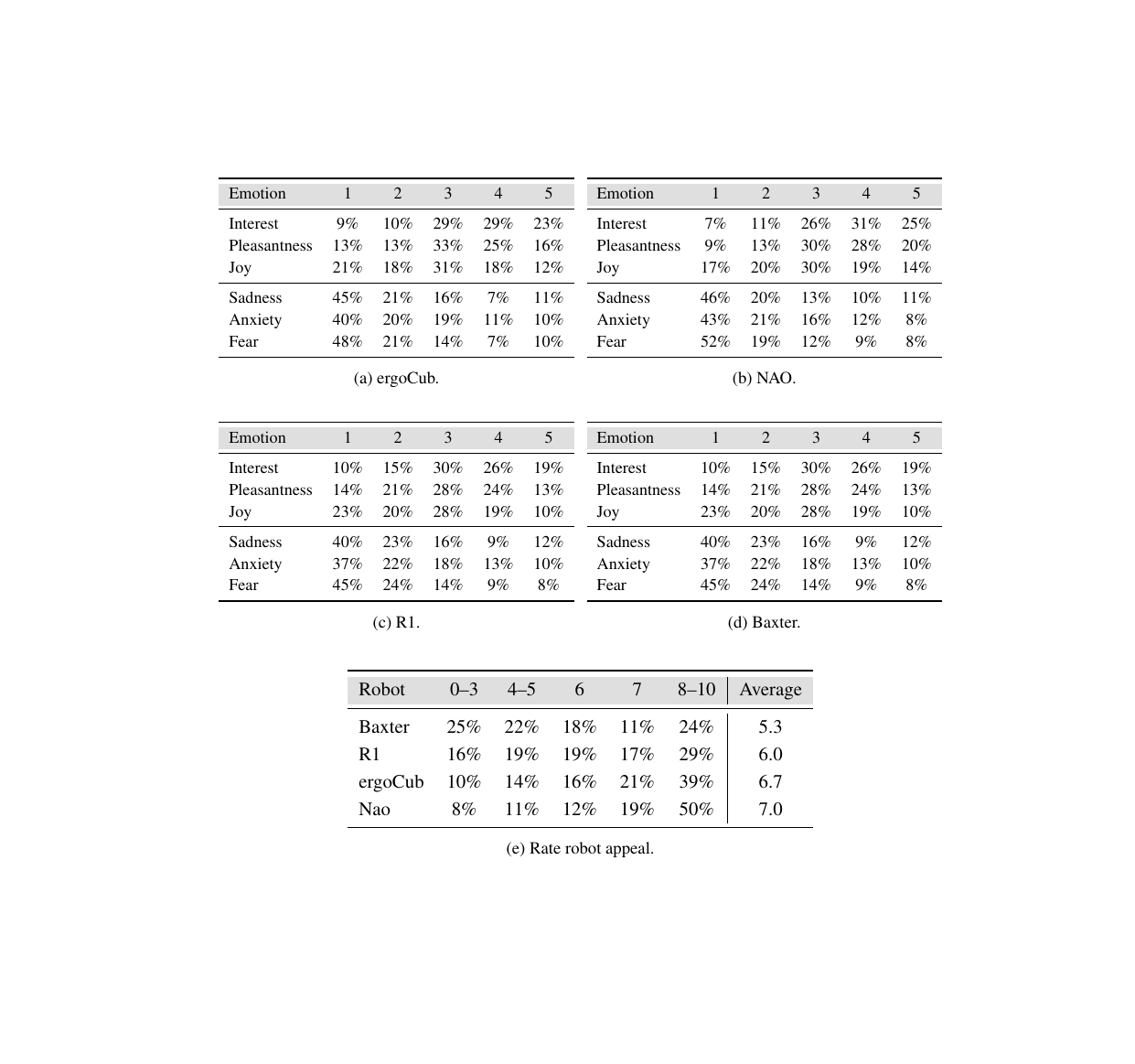}  
  \caption{Comparison of robot acceptability results in the manufacturing domain. Figures (a–d) report the percentage of responses to the question: \textit{“On a scale from 1 (not at all) to 5 (completely), to what extent does the robot image evoke the following sensations?”}. Figure (e) reports the responses to the question: \textit{“On a scale from 1 (not at all) to 10 (completely), how much do you like the robot design?”}.}
  \label{fig:comparison_Manufacture}
\end{figure}
\clearpage
\begin{figure}[!htbp]

        \includegraphics[trim={0.0cm 10cm 0cm 0cm},clip,width=\linewidth]{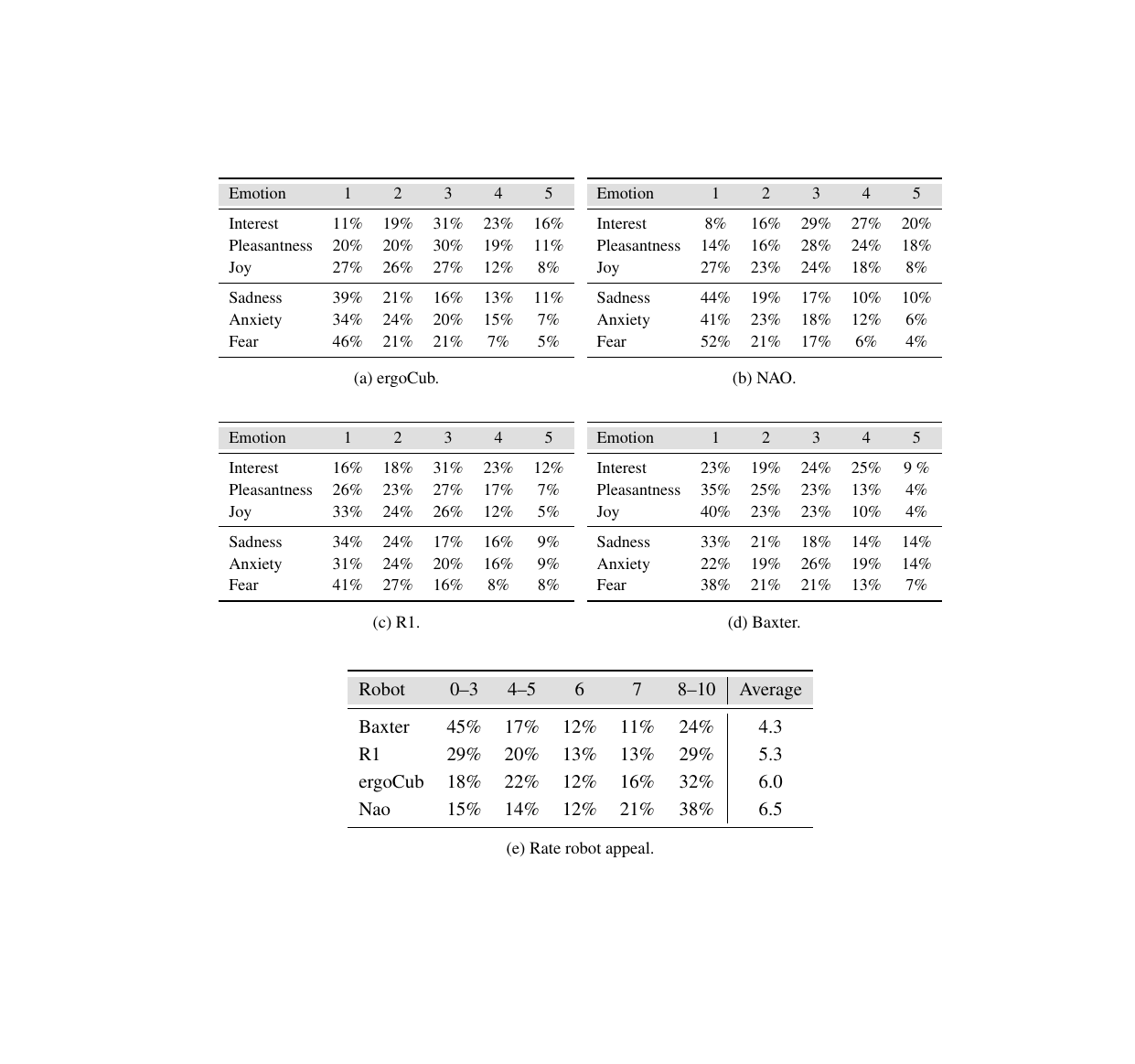}  
  \caption{Comparison of robot acceptability results in the healthcare domain. Figures (a–d) report the percentage of responses to the question: \textit{“On a scale from 1 (not at all) to 5 (completely), to what extent does the robot image evoke the following sensations?”}. Figure (e) reports the responses to the question: \textit{“On a scale from 1 (not at all) to 10 (completely), how much do you like the robot design?”}.}
  \label{fig:comparison_Healtcare}
\end{figure}
\clearpage
\begin{figure}[H]
  
        \includegraphics[trim={0.0cm 14cm 0cm 0cm},clip,width=\linewidth]{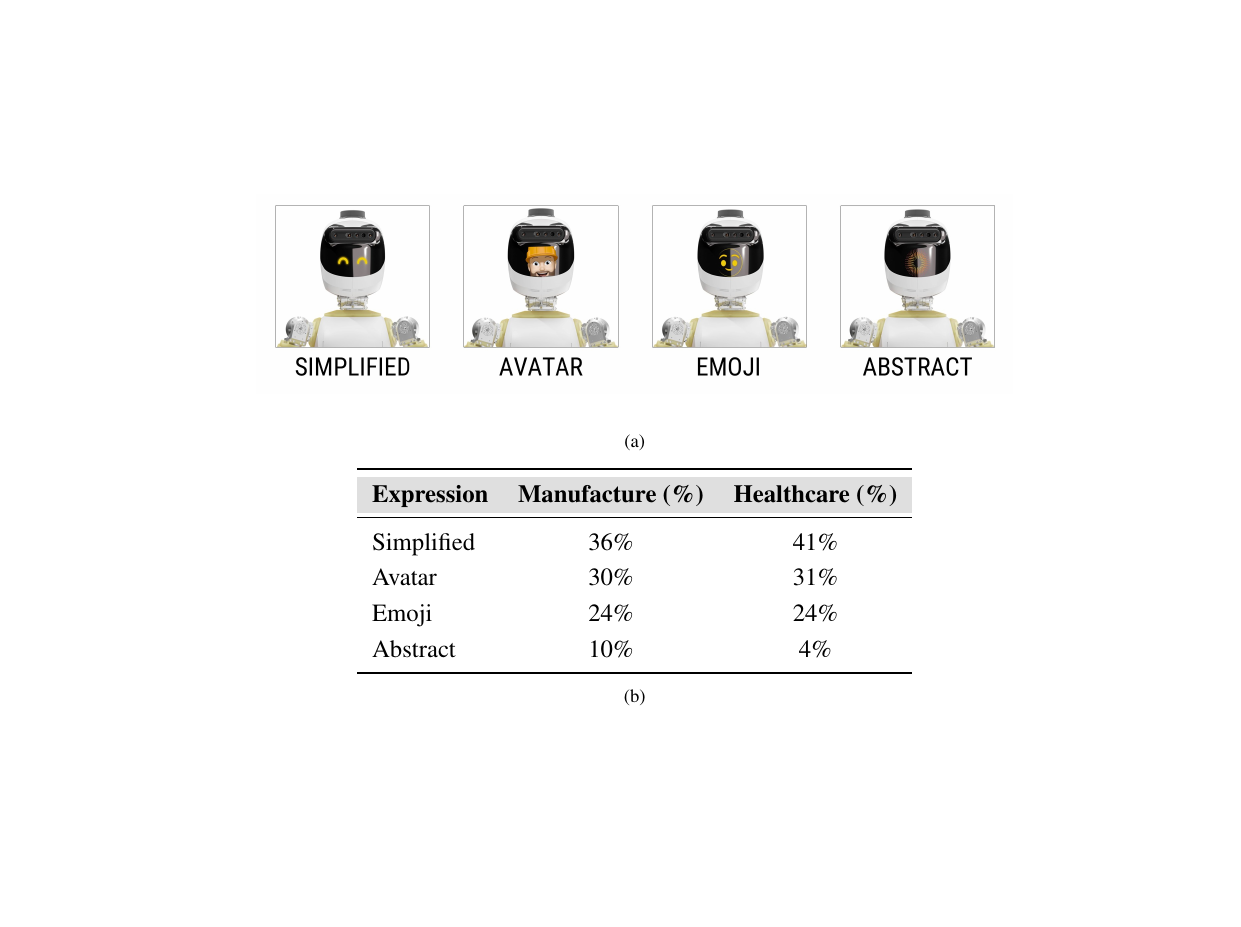}  
  \caption{Acceptability of ergoCub facial expression. (a) The different facial expression considered. (b) Comparison of preferred facial expression styles in the manufacturing and healthcare domains. Participants answered the question “Which facial expression do you prefer?”.}
  \label{fig:comparison_Faces}
\end{figure}

\clearpage
\renewcommand{\thefigure}{E\arabic{figure}}

\setcounter{equation}{0}
\setcounter{figure}{0}
\clearpage
\begin{figure}[t]
        \centering
        \includegraphics[trim={0.0cm 12cm 0cm 0cm},clip,width=\linewidth]{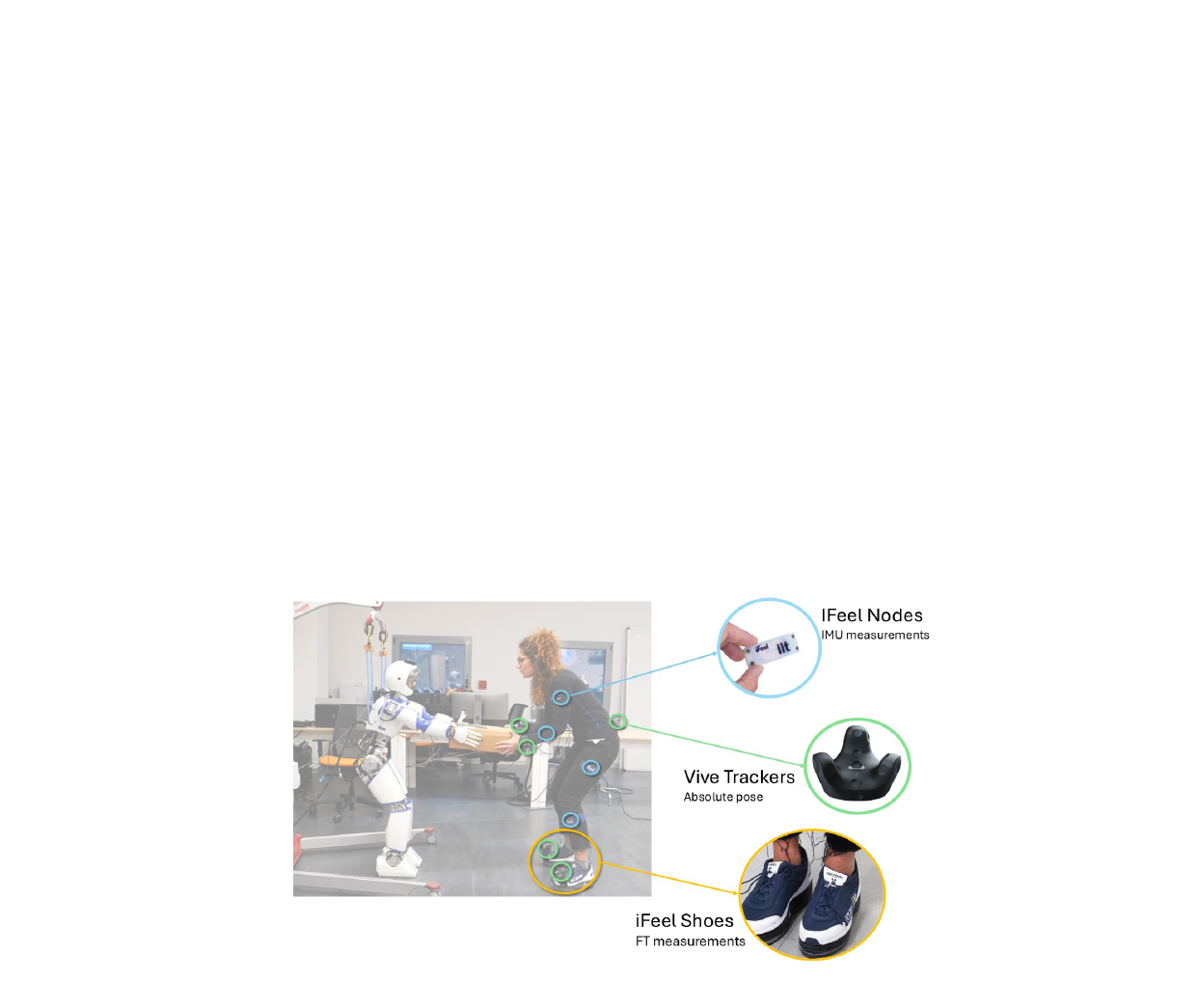}  
        \caption{The human and the robot collaborate to move a load. The figure highlights the non-intrusive sensors worn by the human and the corresponding measurements they provide. The human wears several iFeel Node sensors (blue markers) placed on the arms and legs, which provide IMU measurements. VIVE trackers (green markers) are mounted on the feet (on top of the iFeel Shoes), wrists, and waists to capture the absolute pose of the human body segments. The iFeel Shoes (highlighted in orange) are equipped with force/torque (FT) sensors embedded in the soles to measure ground reaction forces.}
        \label{fig:human_sensors}
    \end{figure}
\clearpage
\newpage
% --- Bibliography inlined from sn-article.bbl ---
%% BioMed_Central_Bib_Style_v1.01

% --- End of bibliography ---
\end{document}